\documentclass{article} 
\usepackage{iclr2024_conference,times}

\usepackage{amsmath,amsfonts,bm}

\def\eqref#1{equation~\ref{#1}}

\def\1{\bm{1}}

\DeclareMathAlphabet{\mathsfit}{\encodingdefault}{\sfdefault}{m}{sl}
\SetMathAlphabet{\mathsfit}{bold}{\encodingdefault}{\sfdefault}{bx}{n}

\def\gF{{\mathcal{F}}}
\def\gG{{\mathcal{G}}}

\def\gJ{{\mathcal{J}}}

\def\gP{{\mathcal{P}}}
\def\gQ{{\mathcal{Q}}}

\def\gT{{\mathcal{T}}}

\def\gW{{\mathcal{W}}}

\usepackage{multirow}
\usepackage{diagbox}
\usepackage{hyperref}
\usepackage{url}
\usepackage{todonotes}
\usepackage{verbatim}
\usepackage{color}  

\definecolor{blue}{rgb}{0, 0, 0}
\def\cblue{\textcolor{blue}}

\usepackage{wrapfig,lipsum,booktabs}

\title{Learning semilinear neural operators: a unified recursive framework for prediction and data assimilation}

\renewcommand\footnotemark{}

\author{
Ashutosh Singh$^{1,*}$ \And Ricardo Borsoi$^{2,*}$ \And Deniz Erdogmus$^1$ \And Tales Imbiriba$^{1,3}$
\thanks{
\hspace{-4.5ex} $*$ denotes equal contribution.
\\[0.025cm]
$^1$ Department of Electrical and Computer Engineering, Northeastern University, Boston MA, 02115, USA.
\\[0.025cm]
$^2$ CNRS, CRAN, Université de Lorraine, Vandoeuvre-lès-Nancy, F-54000 Nancy, France.
\\[0.025cm]
$^3$ Institute for Experiential AI, Northeastern University, Boston MA, 02115, USA.
\\[0.025cm]
\textbf{Emails:} \texttt{\{singh.ashu,d.erdogmus,talesim\}@northeastern.edu}, \texttt{raborsoi@gmail.com}
} 
}

\iclrfinalcopy 
\begin{document}

\maketitle

\vspace{-2ex}

\begin{abstract}

Recent advances in the theory of Neural Operators (NOs) have enabled fast and accurate computation of the solutions to complex systems described by partial differential equations (PDEs).
Despite their great success, current NO-based solutions face important challenges when dealing with spatio-temporal PDEs over long time scales.
Specifically, the current theory of NOs does not present a systematic framework to perform data assimilation and efficiently correct the evolution of PDE solutions over time based on sparsely sampled noisy measurements.
In this paper, we propose a learning-based state-space approach to compute the solution operators to infinite-dimensional semilinear PDEs.
Exploiting the structure of semilinear PDEs and the theory of nonlinear observers in function spaces, we develop a flexible recursive method that allows for both prediction and data assimilation by combining \textit{prediction} and \textit{correction} operations. 
The proposed framework is capable of producing fast and accurate predictions over long time horizons, dealing with irregularly sampled noisy measurements to correct the solution, and benefits from the decoupling between the spatial and temporal dynamics of this class of PDEs.
We show through experiments on the Kuramoto-Sivashinsky, Navier-Stokes and Korteweg-de Vries equations that the proposed model is robust to noise and can leverage arbitrary amounts of measurements to correct its prediction over a long time horizon with little computational overhead.

\end{abstract}

\section{Introduction}

The evolution of many dynamical systems in science and engineering can be described by \emph{partial differential equations} (PDEs), where modeled quantities are often a function of both space and time. Evolving PDEs over time can be computationally very intensive, especially for fine spatiotemporal grids and large scale systems. In this context, great effort has been recently devoted to provide neural-network (NN) approximations of integral operators as solution to such PDEs~\citep{lu2019deeponet,kovachki2021neural, morrill2021neural, kidger2020neural}, namely, \textit{Neural Operators} (NOs), allowing for efficiently approximating the solution to PDE systems.

Despite the efficiency and approximation power of current NOs, the literature lacks a systematic framework for \textit{data assimilation}~\citep{asch2016dataAssimilation} capable of improving the estimation quality and correcting the evolving trajectories based on a small amount of noisy measurements from the system.
This is very important due to advances of sensing technologies and the proliferation of data from dynamical systems in many fields such as Earth surface temperature \citep{jiang2023efficient}, remote sensing imaging \citep{weikmann2021timesen2crop,borsoi2021spectral}, traffic concentration \citep{thodi2023fourier}, fMRI dynamics \citep{singh2021variation, buxton2013physics} and video dynamics \citep{guen2020disentangling}, to name but a few. 
\cite{afshar2022extended} discussed the design of \textit{observers} for the so-called \textit{semilinar} PDEs (a class of common PDEs with a specific form of spatiotemporal dynamics)
and its close connection to extended Kalman filters~\citep{smith1962application, sarkka2023bayesian}. The designed observer is an important tool that can estimate the state of a system governed by some underlying infinite-dimensional PDEs based on measurements, showing a natural connection to data assimilation. However, standard approaches based on Kalman filters and observers require tremendous computational efforts, limiting their application to large-scale systems. 
Recently, a few works aimed at approximating the Kalman filter updates using NNs for  
video prediction~\mbox{\citep{guen2020disentangling},} while the Kalman updates were approximated in~\citep{revach2022kalmannet}. Nevertheless, a general theory and framework for data assimilation exploiting the structure of semilinear PDEs is still missing.

In this paper, we extend the theory of NOs by exploiting the observer design of semilinear PDEs and provide a systematic approach for data assimilation. The resulting method, namely \textit{NO with Data Assimilation} (NODA), is a recursive NO approach that can be used for both estimation, when noisy measurement data is available, and prediction. To overcome the computational challenges inherent to the design and implementation of infinite-dimensional Kalman observers, we propose a data-driven NN approximation leveraging NOs~\citep{li2020fourier} and learning-based Kalman estimators~\citep{guen2020disentangling}. 
Thus, NODA recursively estimates the system's states over time through \textit{prediction} and \textit{update} steps typical to Bayesian filtering approaches but with a small computational cost. We demonstrate through extensive simulations that NODA leads to better prediction performance 
when compared with other closely related NO approaches, while also being able to assimilate data with arbitrary sampling rates.
\cblue{This can have significant impact in practical applications including, e.g., weather and Earth surface temperature forecast~\citep{pathak2022fourcastnet,jiang2023efficient}, remote sensing imaging \citep{weikmann2021timesen2crop,borsoi2021spectral} and fMRI dynamics \citep{singh2021variation,buxton2013physics}.}
The contributions of this work can be summarized as:
\vspace{-0.6cm}
\begin{itemize}
\itemsep-0.03em
    \item we extend the NO theory by leveraging the observer design of semilinear PDEs;
    \item we break the observer solution into prediction and update steps, allowing for a systematic way of performing both prediction and data assimilation, and devise a data-driven solution;
    \item the resulting framework can estimate solutions using arbitrary amounts of measurements. 
\end{itemize}

\vspace{-0.2cm}

\textbf{Related work:}
Traditionally, many methods have been proposed to approximate the solution of PDEs, as can be found in \citep{larson2013finite} and \citep{evans2022partial}. A major drawback of traditional solvers is that they are computationally taxing. A class of these methods resorts to data-driven approaches that aim at approximating the solution operator of the underlying PDE from the snapshots of the states \citep{kutz2016dynamic,krstic2008boundary}. Recently, NO learning has become a popular strategy for learning solution operators to a broad family of parametric PDEs \citep{li2020fourier,li2020neural, kovachki2021neural,brandstetter2022clifford}. These methods have been further used in many applications such as for weather forecast \citep{pathak2022fourcastnet} and to track CO2 migration \citep{wen2022u} and coastal floods \citep{jiang2021digital}, among others. 

NOs are also being adopted in different machine-learning paradigms. \cite{guibas2021adaptive} present work on using a Fourier NO for token mixing in transformers. \cite{goswami2022physics} motivates a physics-informed learning of NOs. \cite{brandstetter2022lie} use the symmetries of Lie groups for data augmentation. \cblue{\cite{gupta2021multiwaveletOperatorLearning} proposed a multiwavelet-based operator leaning framework.} \cite{rotman2023semi} and \cite{kaltenbach2023semi} present a semi-supervised learning approach for NO learning. \cite{hao2023gnot} propose the use of transformers for NO learning. \cite{chen2023laplace} adapts NOs for different geometries. In \citep{magnani2022approximate} the authors provide a Bayesian treatment of NO for uncertainty estimation. 
One of the key drawbacks of these methods is that for long time scales, the solutions tend to deviate from the true trajectory. \cite{li2022learning} present a Markov NO for tracking long-term trajectories of chaotic dissipative systems. However, such works have not considered the presence of noisy measurements, which is common in real-world datasets. This motivates the need for new frameworks that can address data assimilation. 

Similar to the proposed work, \cite{salvi2022neural} attempt to exploit the structure of the solutions of semi-linear stochastic PDEs and propose learning frameworks based on solving a fixed-point problem using ODE numerical solvers. However, the method on \citep{salvi2022neural} differs from the proposed framework as it does not address the need for correcting predictions using noisy measurement available during test time and is not recursive in nature. The capability of learning a solution operator that is flexible, recursive and robust to noisy data is essential for performing fast and accurate predictions and data assimilation over long time horizons.

Data assimilation has been well studied in the context of dynamical systems \citep{cheng2023machine, farchi2021using, levine2022framework, levine2023machine}. Most methods discussed in these works exploit auto-regressive formulations, which can provide good approximations of the dynamical behavior typically seen in such systems.
Recently, \cite{frion2023neuralKoopmanAssimilation} extended deep dynamic mode decomposition to learn Koopman operators through data assimilation. 
In the case of infinite dimensional systems, \cite{afshar2022extended} discusses the observer design of semilinear PDEs and its close relationship with infinite-dimensional extended Kalman filter, which is closely connected to data assimilation methods. We leverage these principles to propose NODA as it will become clear in the remainder of this paper.




\vspace{-0.5ex}

\section{Background}
\label{sec:background}

\vspace{-1ex}

\textbf{Learning Solution Operators to PDEs:}
Many applications ranging from weather forecasting to modelling molecular dynamics involve finding the solution to PDEs in both spatial and temporal variables, herein denoted by $x\in\mathbb{R}^q$ and by $t\in\mathbb{R}$, respectively. The generic PDE considered in recent works is of the form 
\begin{align}
 \begin{aligned}
 (L_a z)(x,t) &= f(x,t) \,, \quad\qquad & (x,t)\in D \,,
 \\
 z(x,t) &= 0 \,, & (x,t)\in \partial D \,.
\end{aligned}
\label{eq:genericPDE}
\end{align}
Here $z\in \mathcal{Z}_{xt}$ represents the solution of the PDE that belongs to a Banach space $\mathcal{Z}_{xt} \subseteq\mathcal{L}(D,\mathbb{R}^p)$, its domain given by a bounded, open set $D\subset \mathbb{R}^q\times\mathbb{R}$ with a Dirichlet boundary condition imposed over $\partial D$, $f \in \mathcal{Z}_{xt}^*$ is a known function, $L_a: \mathcal{Z}_{xt} \rightarrow \mathcal{Z}_{xt}^*$ is a differential operator with a set of parameters $a$, and $\mathcal{Z}_{xt}^*$ is the dual space of $\mathcal{Z}_{xt}$. 
Recent methods in neural operator learning focus on computing a parametric mapping $\gG_{\theta}:a\mapsto z$ which approximates the solution to the PDE in~\eqref{eq:genericPDE} based on a set of pointwise evaluations $\{a_i,z_i\}_{i=1}^N$. Such approximation is performed by minimizing the empirical risk \citep{kovachki2021neural}:
\begin{align}
    \textstyle{
   {\displaystyle \min_{\theta}} \,\, \frac{1}{N} \sum_{i=1}^N \|z_i-\gG_{\theta}(a_i)\|_{\mathcal{Z}}^2 \,,
    }
\end{align}
where $\|\cdot\|_{\mathcal{Z}}$ denotes an appropriate norm on $\mathcal{Z}$. A closely related problem consists in learning an operator that computes a snapshot solution of the PDE at a time $t>0$, $z(\cdot,t)$, based on some initial condition $z(\cdot,0)$ and on a set of parameters $a$.

Different NOs were recently proposed to define the parametric mappings $\gG_{\theta}$, motivated by the kernel integral formulation of the solution operator to linear PDEs~\citep{li2020neural,kovachki2021neural}. These are given in the form of the composition of different \textit{layer} operators, such as:
\begin{align}
    \gG(a) = (\gQ \circ \gW_L \circ \cdots \circ \gW_1 \circ \gP)(a) \,,
\end{align}
with $\gW_{\ell}$ being the operator for layer $\ell$, $\gP$ is a lifting operator, mapping $a$ to the first hidden representation, and $\gQ$ a projection operator, mapping the last hidden representation to the output.
Different forms for the operator layers have been proposed, such as the graph and multipole graph NOs~\citep{kovachki2021neural}, the U-shaped NO~\citep{rahman2023uno}, and the DeepONets~\citep{lu2019deeponet}.

A notable example is the Fourier neural operator (FNO) layer~\citep{li2020fourier}. 
The FNO was motivated by constraining the kernel in the integral operator as a shift-invariant function~\citep{li2020neural}, thereby making the integral operator equivalent to a convolution that can be implemented in the frequency domain. 
An FNO layer, applied to an input signal $v_{\ell}$ and evaluated at $(x,t)\in D$, can be represented generically as
\begin{align}
    v_{\ell+1}(x,t) = \sigma\left(W v_{\ell}(x,t) + \mathcal{F}^{-1}(R_{\phi}\cdot\mathcal{F}(v_{\ell}))(x,t) \right) \,,
    \label{eq:FNO_layer}
\end{align}
where $\sigma$ is a componentwise nonlinear activation function, $W$ is a linear operator, $\gF$ is the Fourier transform, and $R_{\phi}$ is the Fourier transform of the kernel, parametrized by $\phi$. The FNO can be easily discretized for form a NO layer, and be made invariant to the discretization level~\citep{li2020fourier}.




\textbf{Semilinear PDEs:}
Many applications involve PDEs which exhibit a very specific spatiotemporal structure, where the time variable has linear dynamics, while the spatial variables can be governed by more complex and possibly highly nonlinear equations~\citep{curtain2020introduction}. This specific spatiotemporal decoupling, in the so-called \emph{semi-linear form}, is represented as~\citep{rankin1993semilinear}:
%
%
\begin{equation}
    \frac{\partial z(t)}{\partial t} = Az(t) + G(z(t), t) \,, 
    \qquad\qquad
    z(0) = z_0 \,,
    \label{eq:SEEmodel}
\end{equation}
where $z(t) \in \mathcal{Z}_x$ denotes the solution at time index $t\in[0,t_f]$, which is itself a function of the spatial variables $x \in \mathbb{R}^q$, $A : \mathcal{D}(A)\rightarrow \mathcal{Z}_x$ be a linear bounded operator with domain $\mathcal{D}(A)\subseteq \mathcal{Z}_x$ that generates continuous semi-group operator $T(t):\mathcal{Z}_x\to\mathcal{Z}_x$~\citep{curtain2020introduction}, and $G: \mathcal{Z}_x \times[0,t_f]\rightarrow \mathcal{Z}_x$ is an operator which is strongly continuous on $[0,t_f]$ and possibly non-linear in $\mathcal{Z}_x$, satisfying $G(0,t) = 0$. The operator $T(t)$ is closely linked to the temporal evolution of the PDE solution $z(t)$, as will become clearer in the following section.
Note that $z(t)$ is still a function of the spatial variables $x$.
%
Examples of PDEs with this structure include the Navier-Stokes and the Kuramoto-Sivashinsky equations, which can be seen in Section~\ref{sec:experiments}.

\vspace{-1ex}

\section{Prediction and Data Assimilation with Recursive NOs}

\vspace{-1ex}

In this section, we will present the proposed framework for a recursive method to handle both prediction of the PDE solution as well as data assimilation.
Although semilinear PDEs can describe the time evolution of many infinite dimensional systems, in many practical applications such as in weather forecast \citep{pathak2022fourcastnet} or tracking of coastal floods or CO2 migration~\citep{wen2022u,jiang2021digital}, one does not have direct access to measures or discrete snapshots of the true solution $z(t)$. Instead, we only measure a noisy or degraded version thereof in a finite-dimensional Euclidean space, which we denote by $y(t)\in\mathbb{R}^p$. Thus, we consider the following model:
\begin{align}
    \frac{\partial z(t)}{\partial t} &= Az(t) + G(z(t), t) + \omega(t) \,, \label{eq:dynamodel}\\
    z(0) &= z_0 \,, \label{eq:InOutmodel} \\
    y(t) &= Cz(t) + \eta(t) \,. \label{eq:measurement_model}
\end{align}
Note that, besides a dynamical evolution based on a semilinear PDE in equations~(\ref{eq:dynamodel}) and~(\ref{eq:InOutmodel}), the second part of the model in~\eqref{eq:measurement_model} describes how the measurements are generated from the true solution (or \textit{states}) $z(t)$, with $\eta(t) \in \mathbb{R}^p$ representing the measurement noise and $C \in \mathcal{L}(\mathcal{Z}_x, \mathbb{R}^p)$ being a mapping from the solution space to the measurement space. $\omega(t)\in\mathcal{Z}_x$ is an unknown perturbation signal which describes uncertain knowledge of the system dynamics.


As mentioned previously, in many applications we only observe $y(t)$ on a discrete set of (possibly sparse, intermittent) time samples $\gT\subset[0,t_f]$ in an interval of length $t_f$. 
Given an initial condition/initialization $z_{0}$ and the set of discrete measurements $\{y(t_k):t_k\in \gT\}$, our goal is to learn an \textit{online} model capable of recovering/estimating the solution $z(t)$ \textit{recursively} over time, using only present or past measurements $\{y(\tau) : \tau\leq t, \tau\in\gT\}$ at every time $t$. This entails the need for a flexible method, capable of performing \textit{prediction} in order to recover $z(t)$ at time instants when no measurements are available (using only the past estimates of $z(\tau)$ for $\tau<t$, thus dealing with missing data), and to \textit{correct} the solution with the measured data $y(t)$ on the small number of time instants when it is observed, performing data assimilation.

In the following, we will first investigate the particular form of the solutions to the PDE in~(\ref{eq:dynamodel})--(\ref{eq:measurement_model}) and of observer designs that can provide a consistent estimate of its solution $z(t)$. This will be paramount to support the data-driven solution proposed thereafter.

\vspace{-1ex}

\subsection{Observer design for system with semilinear PDEs}

\vspace{-0.5ex}

The specific structure of the semi-linear PDE allows us to write the solution in general form under mild conditions. Thus, assuming that the nonlinear operator $G$ admits a Fréchet derivative that is globally bounded as well as locally Lipschitz, uniformly in time, there exists a solution $z(t) \in \mathcal{Z}_x$ to \eqref{eq:SEEmodel}, for a time interval $0\leq t\leq t_f$, that can be written as \citep{weissler1979semilinear,rankin1993semilinear}
\begin{align}
    z(t) = T(t)z_0 + \int_{0}^{t}T(t-s)G(z(s),s)ds  \,.
    \label{eq:sle_original}
\end{align}
When considering the availability of measurements $y(t)$ as in~\eqref{eq:measurement_model}, these can be used to correct the state evolution of the system by designing an \textit{observer}. This is performed by designing a PDE that contains a copy of the original system’s dynamics with the addition of a correction term which is based on the error between the predicted and actual measurements, hereby denoted by $C\hat{z}(t)-y(t)$, by means of an observer operator gain~\citep{afshar2022extended}.
For equations (\ref{eq:dynamodel})--(\ref{eq:measurement_model}), the general form of the observer is given by
\begin{align}
    \frac{\partial \hat{z}(t)}{\partial t} = A\hat{z}(t) + G(\hat{z}(t), t) +K(t)\big[y(t) - C\hat{z}(t)\big] \,,
    \qquad\qquad 
    \hat{z}(0) = \hat{z}_0 \,,
\label{eq:Observer}
\end{align}
where $K(t):\mathbb{R}^p\to\mathcal{Z}_x$ is the \textit{observer gain}, mapping the error in the measurement space to the PDE dynamics, and $\hat{z}(t)$ is the observer solution.


By solving a nonlinear infinite-dimensional Riccati equation, one can design an operator $K(t)$ based on the PDE in~\eqref{eq:Observer} that can compute the solution $\hat{z}(t)$ over time, such that under the absence of noise and disturbances, and under mild additional conditions, the solution satisfies $\|z(t)-\hat{z}(t)\|_{\mathcal{Z}_x}\to0$ as $t\to\infty$ for sufficiently small initial errors \cblue{(see Theorem~5.1 in \mbox{\citep{afshar2022extended}})}.
\cblue{Furthermore, the authors show that under bounded disturbances $\omega(t)$ and $\eta(t)$, the estimation error can also be bounded for all $t$ \cblue{(see Corollary~5.2 in~\citep{afshar2022extended})}.} 
Similarly to the previous case, \cblue{when $K(t)$ is strongly continuous} the analytical form of a solution $\hat{z}(t)$ to (\ref{eq:Observer}) is given by:
\begin{equation}
    \hat{z}(t) = T(t)\hat{z}_0 + \int_{0}^{t}T(t-s)\Big[G(\hat{z}(s),s) + K(s)[y(s) - C\hat{z}(s)]\Big]ds \,.
    \label{eq:MildSol_Kal}
\end{equation}
The observer solution, however, is challenging to implement. First, the numerical solution to the infinite-dimensional Riccati equation and computation of~\eqref{eq:MildSol_Kal} is difficult and computationally expensive. Moreover, it requires measurements $y(t)$ to be available at all time instants $t\in[0,t_f]$, and cannot handle missing data or perform prediction. These challenges will be addressed in the next section by the learning-based solution.

\vspace{-0.5ex}

\subsection{Learning-based Recursive Prediction-Correction NO}

\vspace{-0.5ex}


In this section, we provide a fully data-driven framework to compute a recursive solution to~(\ref{eq:dynamodel})--(\ref{eq:measurement_model}) which can perform prediction and handle missing data.
First, we discretize the observer solution in the time domain. By rewriting the solution in~\eqref{eq:MildSol_Kal}, we can use it to represent the evolution of the estimate $\hat{z}(t)$ between any two discrete time instants, $t_{k-1}$ and $t_k$ (for convenience, we assume the discretization timesteps $h_k=t_k-t_{k-1}$ to be approximately constant with $k$). 
\cblue{Moreover, it is possible to rewrite this solution in such a form that we can identify two terms, \textit{prediction} and \textit{correction} (see Appendix~\ref{app:derivations_equation12} for details),} leading to:
{\small \begin{align}
\hat{z}(t_k) ={} & \underbrace{T(t_k-t_{k-1})\hat{z}(t_{k-1}) + \int_{t_{k-1}}^{t_k} \!\! T(t_k-s)G(\hat{z}(s),s)ds}_{\rm Prediction}
+ \underbrace{\int_{t_{k-1}}^{t_k} \!\! T(t_k-s)K(s)\big[y(s) - C \hat{z}(s)\big]ds}_{\rm Correction} \,.
\label{eq:sle}
\end{align}}
The prediction term depends only on the past solutions $\hat{z}(t_{k-1})$, predicting the evolution of the PDE states, whereas the correction term is based on the reconstruction error. This structure will be used in the design of the proposed learning-based solution.

Moreover, since most differential operators $A$ generate a semigroup operator $T$ which has an integral form~\citep{gerlach2014semigroups}, this motivates the use of an NO framework to approximate the terms in~\eqref{eq:sle}.
Thus, we propose a learning-based approach that leverages the structure of~(\ref{eq:sle}) to perform both prediction and data assimilation in a flexible manner.
Denote the estimated solution at time $t_k$ and at some spatial discretization level (typically in a regular grid) by $\hat{z}_{t_k}=\Pi_D(\hat{z}(t_k))$, where $\Pi_D$ is a discretization operator which maps from the original solution space $\mathcal{Z}_x$ to some finite-dimensional Euclidean space.
Specifically, we separate the prediction and correction terms in (\ref{eq:sle}), and propose architectures, described in the following, to learn each of these operations separately. The prediction operation is applied at every time instant $t_k$, generating the predicted solution $\hat{z}_{t_k}^{\rm pred}$. Then, to adapt the method to the assimilation setting, if measurements are available (i.e., $y(t_k)$ is observed) the predicted solution is further processed by the correction operation to generate the final estimated solution $\hat{z}_{t_k}$. Otherwise, if there are no measurements, the estimated solution is set as the predicted one as $\hat{z}_{t_k}=\hat{z}_{t_k}^{\rm pred}$. This process, which constitutes the NODA framework, is further described in the following. 



\textbf{Prediction:}
We approximate the first term in~\eqref{eq:sle}, which \textit{predicts} of the solution at time $t_k$ given the estimate at time $t_{k-1}$, with a learnable NO in a residual formulation, given by:
\begin{align}
    \hat{z}_{t_k}^{\rm pred} &= \hat{z}_{t_{k-1}} + 
    \gW(\hat{z}_{t_{k-1}}) \,,
    \label{eq:predmodel}
\end{align}
where $\gW$ is a neural operator.
Note that this operation inherits the discretization invariance from the NO $\gW$, such as in the FNO~\citep{li2020fourier}. In this work, we use as $\gW$ an FNO layer as defined in~\eqref{eq:FNO_layer}, with the Fourier transform computed only over the spatial dimension, $x$. 
If there are no measurements available for assimilation at time $t_k$, then we set the estimated solution as the predicted one, $\hat{z}_{t_k}=\hat{z}_{t_k}^{\rm pred}$. If a measurement is available, however, the predicted solution is further added to the correction term as described in the following.



\textbf{Correction:}
We approximate the second term in~\eqref{eq:sle}, which \textit{corrects} the predicted solution $\hat{z}_{t_k}^{\rm pred}$ based on the available measurement $y(t_k)$, using the following model, inspired from the observer operator form in~(\ref{eq:sle}) as
\begin{align}
    \hat{z}_{t_k} &= \hat{z}_{t_k}^{\rm pred} + K(\hat{z}_{t_k}^{\rm pred}) \big[y(t_k) - E(\hat{z}_{t_k}^{\rm pred})\big] \,,
    \label{eq:corrmodel}
\end{align}
where $E$ is a learnable operator which approximates the (possibly unknown) measurement operator $C$ in \eqref{eq:measurement_model} over a given discretization level.
In this work, we parametrized $E$ using a one-hidden layer fully connected ReLU NN, which can also tackle cases where the measurement model might be nonlinear.

\cblue{To parametrize the observer gain $K$, we generalized the gating architecture used in~\citep{guen2020disentangling} to cope with general measurement operators $C$ and the observer design:}
\begin{align}
    K(\hat{z}_{t_k}^{\rm pred})[u] &= \mathrm{tanh}\big(W_z E(\hat{z}_{t_k}^{\rm pred}) + W_y y(t_k) + b\big) \,\odot \, \big(\hat{C}^*u\big) \,,
    \label{eq:obsgain}
\end{align}
where $\odot$ denotes the Hadamard product and $W_z$, $W_y$ and $b$ are learnable linear operators. $\hat{C}^*$ is the conjugate operator of $\hat{C}=C\circ \Pi_D$, $\hat{C}:\hat{z}_t\mapsto y(t)$ being the discretized version of $C$. Note, however, that $\hat{C}^*$ can be a learnable linear operator in~(\ref{eq:obsgain}), particularly when $C$ is unknown.


\textbf{Learning criterion:}
Given a set of training data with $S$ different realizations of trajectories of discretized solutions to the PDE (\ref{eq:dynamodel})--(\ref{eq:measurement_model}), each generated from a different random initial conditions $z_D^{(i)}(t_0)$ and containing $N$ time samples, which we denote by $\{z_D^{(i)}(t_k),y^{(i)}(t_k)\}_{k=1}^N$, for $i=1,\ldots,S$, NODA is learned by minimizing the following loss function:
\begin{align}
    \gJ(\phi) = \frac{1}{SN} \sum_{i=1}^S \sum_{k=1}^{N} \big\|\hat{z}_{t_k}^{(i)} - z_D^{(i)}(t_k)\big\|_{2} +  \frac{\lambda}{SH} \sum_{i=1}^S \sum_{k=1}^{H} \big\|y^{(i)}(t_k) - E\big(\hat{z}_{t_k}^{(i)}\big)\big\|_{2} \,,
\end{align}
where $z_D^{(i)}(t_k)=\Pi_D (z^{(i)}(t_k))$ denotes the the discretization of the PDE trajectory, and $\hat{z}_{t_k}^{(i)}$ denotes the recovered solution to the PDE, for the $i$-th realization. To address both the prediction and data assimilation objectives during training, NODA is initialized with $\hat{z}_{t_0}^{(i)}=z_D^{(i)}(t_0)$, then, we supply it with measurements up to a time index $t_H\leq t_N$, i.e., $\{y^{(i)}(t_k)\}_{k=1}^H$, which are used to perform data assimilation in the interval $[t_1,t_H]$, after which it performs only predictions between $t_{H+1}$ and $t_N$, to generate the estimated trajectory $\{\hat{z}_{t_k}^{(i)}\}_{k=1}^N$.
The loss function $\gJ(\phi)$ contains two terms. The first term measures the reconstruction of the PDE solution, while the second term measures the ability of the model to reconstruct the measurements, and the balance between these two terms is controlled by the weight $\lambda\geq0$. The second term is particularly important when learning the operator $E$. 
The learnable parameters are denoted by $\phi$, and contains the parameters needed for the prediction (i.e., the NO $\gW$) as well as the correction step (i.e., $E$ and the parameters of $K$).

\section{Experiments}
\label{sec:experiments}


In this section, we investigate the use of the proposed NODA framework on 1D and 2D semilinear PDEs. We present experiments with the intent of showing the efficacy of the proposed model for:
$\textbf{a)}$ Accurately predicting trajectories over long temporal horizons using only prediction steps; 
$\textbf{b)}$ The efficacy of the correction step and the influence of measurement noise in the estimation performance; and 
$\textbf{c)}$ The influence of the amount of sparsely measured data in the assimilation performance.
For \textbf{b)} and \textbf{c)}, the snapshot measurements $y(t_k)$ are generated by synthetically adding white Gaussian noise to the clean trajectories $z_D(t_k)$, with signal-to-noise (SNR) ratios varying between 10dB and 30dB. Moreover, the noiseless case is also considered in some examples. For the third experiment, \textbf{c)}, we define as $\alpha\in[0,1]$ the ratio between the amount of available (noisy) measurement data (sampled at random instants) and the length of the prediction horizon $\tau_{\rm p} = [t_{H+1},t_f]$. 
To exploit the potential of NODA during test time, a \textit{warm-up} procedure where prediction and update steps are performed at every time step. We define the warm-up horizon window as $\tau_{\rm{w}} = [t_1,t_H]$, and assume that data is available for every time step $t_k\in\tau_{\rm w}$. In all experiments, we evaluate NODA's predictive performance over the prediction horizon $\tau_{\rm p}$ for different values of $\alpha\%$ (including $\alpha=0\%$ to evaluate only the prediction performance), of $t_f$, and for different SNRs, depending on each experiment. All the competing algorithms are initialized with the true solution after the warm-up period, $z_D(t_{H-1})$.




We benchmark our results on a 2D PDE for the Navier-Stokes (NS) equation and on a 1D PDE for the Kuramoto-Sivashinsky (KS) equation. We also performed experiments on the Kortweg-De Vries (KdV) equation~\citep{wazwaz2010partial}, but due to space limitations the results are only included in Appendix~\ref{app:KSresults}.
These data sets represent a family of chaotic semi-linear PDEs. 

\begin{table}[!t]
\footnotesize
\centering
\renewcommand{\arraystretch}{1.2}
\caption{Averaged RelMSE \cblue{($\times 10^3$)} for prediction ($\alpha=0\%$) on the Kuramoto-Sivashinsky equation as a function of the sequence length $t_f$ and of the SNR.}

\setlength{\tabcolsep}{5.5pt}

\fontsize{8pt}{8pt}\selectfont
\begin{tabular}{l|rrr|rrr|rrr}
\hline
 SNR     &    \multicolumn{3}{c|}{$\rm 20$ dB} &       \multicolumn{3}{c|}{$\rm 30$ dB}         &   \multicolumn{3}{c}{$\rm \infty$}    \\ \hline 
 $t_f$  & \multicolumn{1}{c}{60}  & \multicolumn{1}{c}{90} & \multicolumn{1}{c|}{120}  & \multicolumn{1}{c}{60}  & \multicolumn{1}{c}{90}    & \multicolumn{1}{c|}{120} & \multicolumn{1}{c}{60}  & \multicolumn{1}{c}{90}  & \multicolumn{1}{c}{120}  \\ [0.03cm] \hline





MNO		& 144$\pm 7$	&	403$\pm12$	&	543$\pm16$	&	108$\pm6$	&	379$\pm15$	&	521$\pm16$	&	\textbf{13}$\pm1$	&	204$\pm8$	&	410$\pm17$	\\	
FNO		&	298$\pm9$	&	412$\pm12$	&	715$\pm14$	&	259$\pm10$	&	324$\pm10$	&	659$\pm13$	&	223$\pm11$	&	643$\pm19$	&	892$\pm18$	\\	
LSTM		&	303$\pm15$	&	730$\pm15$	&	1.090$\pm33$	&	261$\pm13$	&	483$\pm10$	&	981$\pm10$	&	279$\pm14$	&	688$\pm21$	&	953$\pm19$	\\

NODA		&	\textbf{123}$\pm10$	&	\textbf{329}$\pm13$	&	\textbf{493}$\pm10$	&	\textbf{63}$\pm9$	&	\textbf{310}$\pm9$	&	\textbf{483}$\pm10$	&	19$\pm4$	&	\textbf{185}$\pm11$	&	\textbf{405}$\pm12$	\\	\hline

\end{tabular}
\label{tab:prediction_KS_1}
%
%
\vspace{0.5ex}
%
\caption{Averaged RelMSE \cblue{($\times10^3$)} for prediction ($\alpha=0\%$) on the Navier-Stokes equation as a function of the sequence length $t_f$ and of the SNR.} 
\renewcommand{\arraystretch}{1.2}

\fontsize{8pt}{8pt}\selectfont

\begin{tabular}{l|rrr|rrr|rrr}
\hline
SNR  & \multicolumn{3}{c|}{$\rm 20$ dB} &                                                      \multicolumn{3}{c|}{$\rm 30$ dB}    &    \multicolumn{3}{c}{$\rm \infty$}  \\ \hline
 $t_f$                                             & \multicolumn{1}{c}{500}           & \multicolumn{1}{c}{750}               & \multicolumn{1}{c|}{1000}         & \multicolumn{1}{c}{500}          & \multicolumn{1}{c}{750}          & \multicolumn{1}{c|}{1000}         & \multicolumn{1}{c}{500}          & \multicolumn{1}{c}{750}          & \multicolumn{1}{c}{1000}         \\[0.03cm] \hline

MNO 	&	48$\pm7$	&	72$\pm6$	&	96$\pm12$	&	39$\pm4$	&	51$\pm7$	&	87$\pm6$	&	18$\pm3$	&	22$\pm3$	&	26$\pm5$	\\	
FNO		&	192$\pm10$	&	294$\pm12$	&	351$\pm14$	&	189$\pm19$	&	265$\pm21$	&	317$\pm16$	&	136$\pm11$	&	177$\pm14$	&	265$\pm16$	\\	
\cblue{MWNO}	&	\cblue{172$\pm10$}	&	\cblue{226$\pm9$}	&	\cblue{278$\pm11$}	&	\cblue{147$\pm9$}	&	\cblue{192$\pm9$}	&	\cblue{240$\pm10$}	&	\cblue{78$\pm5$}	&	\cblue{104$\pm6$}	&	\cblue{188$\pm11$}	\\
C-LSTM		&	487$\pm15$	&	522$\pm10$	&	604$\pm18$	&	397$\pm20$	&	426$\pm17$	&	481$\pm19$	&	366$\pm11$	&	417$\pm13$	&	529$\pm11$	\\	
NODA		&	\textbf{10$\pm2$}	&	\textbf{18$\pm4$}	&	\textbf{32$\pm6$}	&	\textbf{8$\pm1$}	&	\textbf{13$\pm2$}	&	\textbf{26$\pm4$}	&	\textbf{7$\pm1$}	&	\textbf{8$\pm1$}	&	\textbf{13$\pm2$}	\\
\hline

\end{tabular}
\label{tab:prediction_NS_1}
%
%
\vspace{0.5ex}
%
\renewcommand{\arraystretch}{1.2}
\caption{Averaged RelMSE \cblue{($\times10^3$)} of NODA for the Navier-Stokes equation as a function of $\alpha$.} 

\fontsize{9pt}{9pt}\selectfont

\begin{tabular}{c|cccc}
\hline
$\alpha$ & 0\% & 10\% & 20\% & 30\% \\ \hline


Averaged RelMSE &	26$\pm4$	&	18$\pm4$	&	13$\pm3$	&	9$\pm3$	\\\hline

\end{tabular}

\vspace{-1ex}

\label{tab:NS_versus_alpha}
\end{table}

\textbf{Benchmark Methods:} We compare NODA with the NO-based methods \textbf{FNO}~\citep{li2020fourier}, \textbf{MNO}~\citep{li2022learning}, \cblue{and \textbf{MWNO}~\citep{gupta2021multiwaveletOperatorLearning} (for the NS),} and with the auto-regressive neural network architectures \textbf{Conv-LSTM}~\citep{shi2015convolutional} (for the NS) and \textbf{LSTM}~\citep{chung2014empirical} (for the KS). \cblue{The MNO and LSTM-based methods were used as baselines due to their recurrent nature, which is also one of the main aspects of NODA, while FNO was selected since it is one of the main building blocks used in the NODA architecture.} To evaluate the accuracy of the results of the different methods, we use the average Relative Mean Squared Error (RelMSE) to compare the accuracy of the trajectories estimated by each method, which is defined as ${\rm RelMSE} = {\mathbb{E}\big\{\sum_{t_k={t_H}}^{t_f}\|z_D(t_k) - \hat{z}(t_k)\|^2_2 \big/ \sum_{t_k={t_H}}^{t_f} \|z_D(t_k)\|^2_2 \big\}}$, 
where $z_D(t_k)$ is the discretized ground truth sequence, $\hat{z}(t_k)$ is the sequence estimated by the method, and $\mathbb{E}\{\cdot\}$ denotes the expected value operator, which is computed over the random initializations that generate the trajectories $z_D(t_k)$.
%
Details on how the loss function is optimized, and on architecture and hyperparameter selections are given in Appendix~\ref{app:implementationdetails}.

\vspace{-1.5ex}

\subsection{Datasets and setup}




\textbf{Kuramoto-Sivashinsky (KS) equation:} The one dimensional KS equation can be represented as $\frac{\partial z}{\partial t} = -z\frac{\partial z}{\partial x}-\frac{\partial^2 z}{\partial x^2} - \frac{\partial^4 z}{\partial x^4}$,
where $z(t)$ is defined over the spatial domain $x \in [0, J]$, with initial condition $z(0) = z_0$. For the experiments we selected $J = 64\pi$ and periodic boundary conditions on $[0,J]$.
We generate 1200 sequences with $t_f\in\{60,90,120\}$ seconds each and a timestep of $h=0.25$ seconds from independent initializations are generated by randomly sampling $z_0$ according to the procedure described in~\citep{li2022learning}
We used 1000 sequences for training the different methods, and the remaining 200 for the evaluation of their performance. The resolution was fixed at 512 samples.
To test NODA we used a warm-up period of $t_H=40$ seconds.


\textbf{Navier-Stokes (NS) Equation:} The 2D Navier-Stokes equation represents a viscous, incompressible fluid in vorticity form on the unit torus, which can be represented as
$\frac{\partial z(x,t)}{\partial t} + u(x,t)\nabla z(x,t) = v\Delta z(x,t) +f(x)$ and $\nabla \cdot u(x,t)=0$, where $u$ is the velocity
field, $z=\nabla \times u$ is the vorticity, $v = \frac{1}{Re}$, $Re > 0$ is the Reynolds number, 
the forcing term is $f(x) = \sin(2\pi(x+y)) + \cos(2\pi(x+y))$, and the initial condition is $z(x,0)=z_0(x)$.
\cblue{Note that the NS equation can be written in semilinear form (see Section~3.3 of~\citep{temam1982behaviourSemilinear_t0}).}
%
\begin{wrapfigure}[18]{r}{8.5cm}
    \centering
    \vspace{-0.0cm}
    \includegraphics[width=0.9\linewidth]{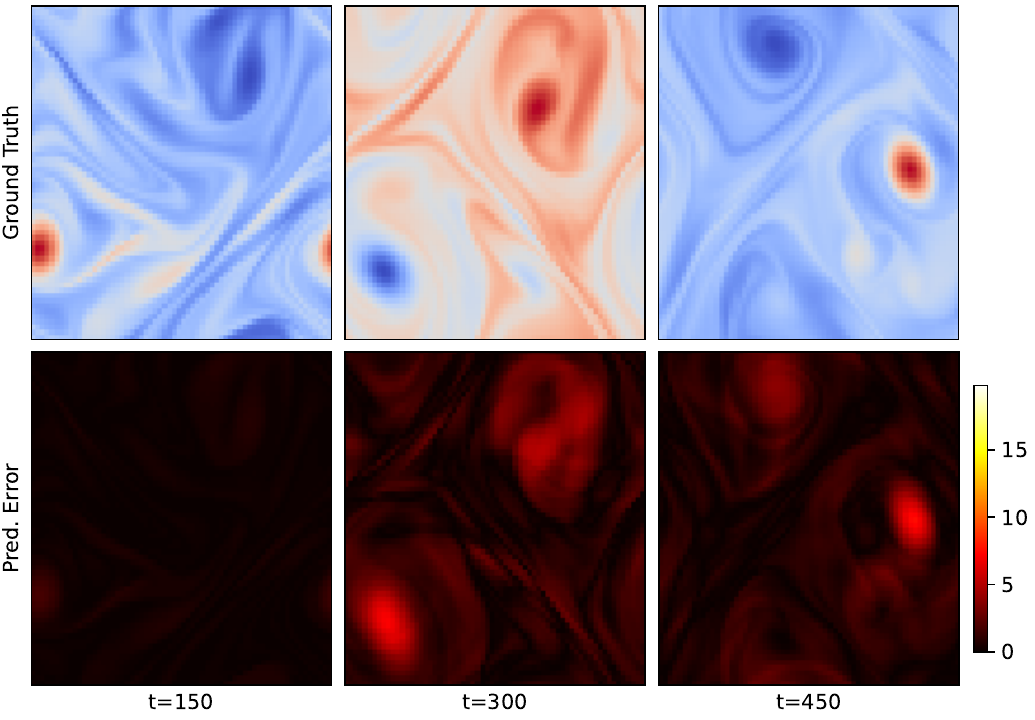}
    \vspace{-0.35cm}
    \caption{Samples of a realization of a true trajectory of the Navier-Stokes equation for $t \in \{150,300,450\}$ (\textbf{Top Row}), and elementwise error plots for the corresponding predictions by NODA (\textbf{Bottom Row}).}
    \label{fig:NS_pred_err}
\end{wrapfigure}
%
%
%
We generated 200 independent trajectories with $t_f\in\{500,750,1000\}$ seconds each and a timestep of $h=1$ second by randomly sampling $z_0$ according to the procedure described in~\citep{li2022learning}.
We use 180 trajectories for training and evaluate on the remaining 20. We fix the resolution at 64$\times$64 for both training and testing. We select $Re = 40$, which presents a non-turbulent case for NS, for benchmarking and comparisons. To test NODA we used a warm-up period of $t_H=50$ seconds.
\cblue{Additional experiments for with higher Reynolds numbers ($Re\in\{500,5000\}$) are available in Appendix~\ref{app:NS_more_vorticities}.}

\subsection{Results and discussion}

\textbf{a) Prediction performance:} We evaluate the prediction of the solutions over the test interval $[t_{H+1},t_f]$ (excluding the samples that were used as warm-up). We compare NODA with no data assimilation (i.e., using $\alpha=0\%$) to the remaining methods, for different sequence lengths $t_f$, SNRs, and for the KS and NS equations (additional results for the KdV equation can be found in Appendix~\ref{app:KSresults}).
We compute the averaged RelMSE results, 
which is presented for the KS and NS equations in Tables~\ref{tab:prediction_KS_1} and~\ref{tab:prediction_NS_1}, respectively.

\begin{figure}[htb]
\centering
\includegraphics[trim={0 0 0 0.1cm},clip,width=0.7\linewidth]{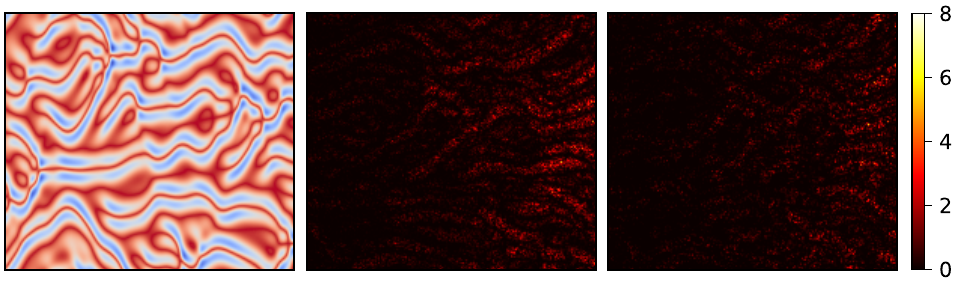}
    \vspace{-0.4cm}
    \caption{Prediction error for one realization of the KS equation. Warm-up with $t_H=40$ was performed. Plots depict the prediction period $(40,200]$ seconds. Ground truth evolution (\textbf{Left}). Error plots for NODA's solutions with $\alpha=0\%$ (\textbf{Center}) and $\alpha = 30\%$ (\textbf{Right}).}
    \label{fig:KS_GT_errors}
\end{figure}

It can be seen that the averaged errors obtained by NODA are in general significantly lower than those achieved by the competing methods, with the MNO being the second-best performing approach, followed by the \cblue{MWNO (for the NS example), by the} FNO and by the LSTM-based methods. This behavior is consistent across the two equations, although MNO achieves slightly better results for $\rm SNR=10dB$ and $t_f=60$ in the KS equation. However, we note that MNO uses the Sobolev norm and dissipativity regularizations in the training objective instead of the Euclidean norm as in the other methods. This illustrates the ability of NODA to outperform competing methods on both clean and noisy data as well as short and long rollouts.
%
A visualization of samples of the ground truth trajectory, as well as error plots for the NODA method and the NS equation can be seen in Figure~\ref{fig:NS_pred_err}. The plots demonstrate the high accuracy obtained with NODA predictions, which errors that show no considerable increase between $t=300$ and $t=450$.
Prediction results for the NS and KS equations are shown in Appendix~\ref{app:sec:moreresults}.

We also evaluate the influence of the amount of samples $t_H$ used in the warm-up step of NODA on its predicting performance by computing the averaged RelMSE as a function of $t_H$ for the NS equation for $\alpha=0\%$, $t_f=1000$ and two different SNRs. It can be seen that the error decreases considerably between $t_H=1$ and $t_H=150$, after which is stabilizes. This shows the benefit of the warm-up step in the performance of NODA, and also demonstrates that relatively small values of $t_H$ suffice to provide considerable performance improvements. Moreover, comparing these plots to the results the other NO-based approaches (i.e., MNO and FNO), we see that NODA provides competitive performance even for $t_H=1$ (i.e., performing predictions based on a single data snapshot), particularly for noisy data, and significantly better for larger values of $t_H$.
%

\begin{figure}[htp]

\centering
\includegraphics[width=.45\textwidth]{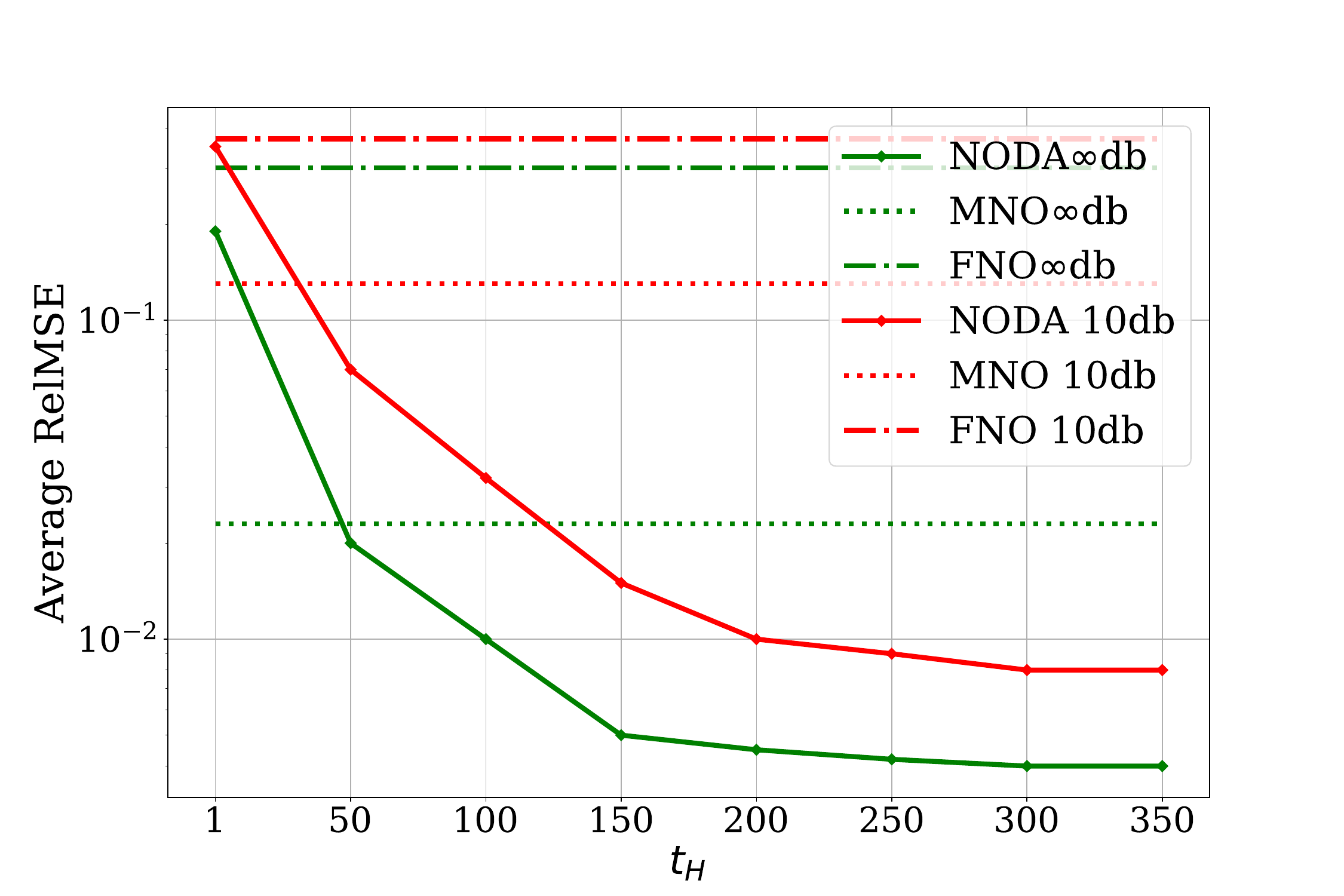}\hfill
\includegraphics[width=.45\textwidth]{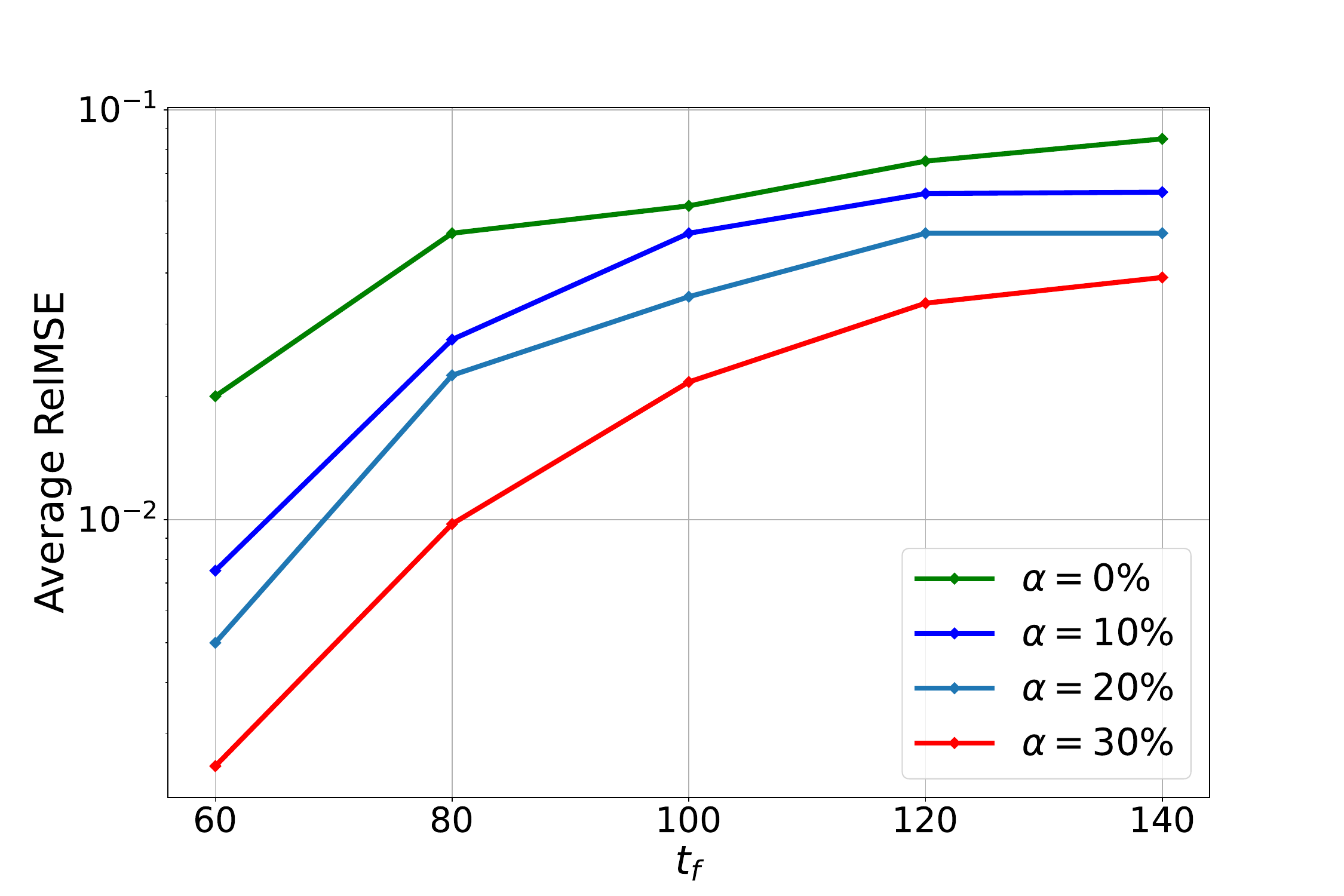}\hfill
\vspace{-0.3cm}
\caption{Average RelMSE as a function of the warm-up length $t_H$ for the NS equation for different SNRs (\textbf{Left}); and average RelMSE as function of $t_f$ for different values of $\alpha$ for the \cblue{KS} equation, where a warm-up period of $t_H=40$ being used for NODA (\textbf{Right}).}
\label{fig:figure3}
\end{figure}

\textbf{b) Data assimilation performance:}
We evaluate the performance of NODA while performing data assimilation in the test interval $[t_{H+1},t_f]$ (i.e., when $\alpha>0$). We evaluate the averaged RelMSE for the NS equation with $\rm SNR=30dB$ and $t_f=1000$ with different data assimilation sampling rates $\alpha\in\{0\%, 10\%, 20\%, 30\%\}$. The results are shown in Table~\ref{tab:NS_versus_alpha}, from which it can be seen that the data assimilation provides an important improvement to the estimation performance of NODA, which increases consistently with $\alpha$. 
Moreover, in the right panel of Figure~\ref{fig:figure3} we also show the averaged RelMSE for the KS equation as a function of the both the sequence length $t_f\in[60,140]$ and of $\alpha$, for an $\rm SNR=10dB$. It can be seen that the error increases only moderately with the length of the prediction interval for all tested values of $\alpha$, which indicates that NODA performs well when estimating both short and long trajectories. Moreover, the averaged RelMSE decreases consistently with $\alpha$, demonstrating the improvements of data assimilation on the results.

A realization of the ground truth and elementwise prediction errors of NODA for the KS equation, both without data assimilation and for $\alpha=30\%$ is shown in Figure~\ref{fig:KS_GT_errors}. It can be seen that the prediction errors tend to increase moderately with $t$, which is consistent with what was observed in Figure~\ref{fig:figure3}. Moreover, using data assimilation with NODA considerably reduces the prediction errors compared to the case where $\alpha=0\%$. Additional results showcasing the data assimilation performance of NODA on the KdV equation are shown in Appendix~\ref{app:KSresults}.

\cblue{\textbf{c) Computation overhead:}
To assess NODA's computational overhead in comparison with other NO-based algorithms, we measured the average time required for the prediction plus assimilation per sample for the NS ($Re=40, \alpha = 30\%)$ example.
The fastest algorithm was the MNO, which took $8\times10^{-5}$ seconds. FNO was the slowest method, taking $2.25$ times as long as MNO, while NODA had an intermediate performance, taking $1.5$ times as long as MNO, illustrating its small computational overhead compared to MNO.}

\section{Conclusions}

In this work, we proposed NODA, a recursive neural operator framework for prediction and data assimilation. To do so, we extended the NO concepts by leveraging the structure of semilinear PDE systems and the correction-based state estimation paradigm from the theory of infinite dimensional observer design. As a result, NODA incorporates prediction and update steps, allowing it to perform both prediction and data assimilation depending on the availability of noisy measurement data. Extensive experiments demonstrate the capability of the proposed methodology to provide significantly more accurate estimations of the solutions of the systems even when only prediction is performed, and especially when doing data assimilation. We also show that even without the warm-up phase (i.e., when performing predictions using only a single snapshot of data), NODA solutions are still far better than the ones provided by its direct counterpart, the FNO. Nevertheless, NODA's framework is flexible and can be directly extended to incorporate other NOs and training losses proposed in the literature, such as the approach adopted by the MNO.


\subsubsection*{Acknowledgments}
\cblue{This work was supported in part by the French National Research Agency, under grants ANR-23-CE23-0024, ANR-23-CE94-0001, and by the National Science Foundation, under grant NSF 2316420, and the National Institute of Health, NIH OT2OD030524.
}





\bibliography{iclr2024_conference}

\begin{thebibliography}{54}
\providecommand{\natexlab}[1]{#1}
\providecommand{\url}[1]{\texttt{#1}}
\expandafter\ifx\csname urlstyle\endcsname\relax
  \providecommand{\doi}[1]{doi: #1}\else
  \providecommand{\doi}{doi: \begingroup \urlstyle{rm}\Url}\fi

\bibitem[Afshar et~al.(2023)Afshar, Germ, and Morris]{afshar2022extended}
Sepideh Afshar, Fabian Germ, and Kirsten Morris.
\newblock Extended {Kalman} filter based observer design for semilinear
  infinite-dimensional systems.
\newblock \emph{IEEE Transactions on Automatic Control}, 2023.

\bibitem[Asch et~al.(2016)Asch, Bocquet, and Nodet]{asch2016dataAssimilation}
Mark Asch, Marc Bocquet, and Ma{\"e}lle Nodet.
\newblock \emph{Data assimilation: methods, algorithms, and applications}.
\newblock SIAM, 2016.

\bibitem[Borsoi et~al.(2021)Borsoi, Imbiriba, Bermudez, Richard, Chanussot,
  Drumetz, Tourneret, Zare, and Jutten]{borsoi2021spectral}
Ricardo~Augusto Borsoi, Tales Imbiriba, Jos{\'e} Carlos~Moreira Bermudez,
  C{\'e}dric Richard, Jocelyn Chanussot, Lucas Drumetz, Jean-Yves Tourneret,
  Alina Zare, and Christian Jutten.
\newblock Spectral variability in hyperspectral data unmixing: A comprehensive
  review.
\newblock \emph{IEEE geoscience and remote sensing magazine}, 9\penalty0
  (4):\penalty0 223--270, 2021.

\bibitem[Brandstetter et~al.(2022{\natexlab{a}})Brandstetter, Berg, Welling,
  and Gupta]{brandstetter2022clifford}
Johannes Brandstetter, Rianne van~den Berg, Max Welling, and Jayesh~K Gupta.
\newblock Clifford neural layers for {PDE} modeling.
\newblock \emph{arXiv preprint arXiv:2209.04934}, 2022{\natexlab{a}}.

\bibitem[Brandstetter et~al.(2022{\natexlab{b}})Brandstetter, Welling, and
  Worrall]{brandstetter2022lie}
Johannes Brandstetter, Max Welling, and Daniel~E Worrall.
\newblock Lie point symmetry data augmentation for neural {PDE} solvers.
\newblock In \emph{International Conference on Machine Learning}, pp.\
  2241--2256. PMLR, 2022{\natexlab{b}}.

\bibitem[Buxton(2013)]{buxton2013physics}
Richard Buxton.
\newblock The physics of functional magnetic resonance imaging {(fMRI)}.
\newblock \emph{Reports on Progress in Physics}, 76\penalty0 (9):\penalty0
  096601, 2013.

\bibitem[Chen et~al.(2023)Chen, Liu, Li, Meng, and Chen]{chen2023laplace}
Gengxiang Chen, Xu~Liu, Yingguang Li, Qinglu Meng, and Lu~Chen.
\newblock Laplace neural operator for complex geometries.
\newblock \emph{arXiv preprint arXiv:2302.08166}, 2023.

\bibitem[Cheng et~al.(2023)Cheng, Quilodr{\'a}n-Casas, Ouala, Farchi, Liu,
  Tandeo, Fablet, Lucor, Iooss, Brajard, et~al.]{cheng2023machine}
Sibo Cheng, C{\'e}sar Quilodr{\'a}n-Casas, Said Ouala, Alban Farchi, Che Liu,
  Pierre Tandeo, Ronan Fablet, Didier Lucor, Bertrand Iooss, Julien Brajard,
  et~al.
\newblock Machine learning with data assimilation and uncertainty
  quantification for dynamical systems: a review.
\newblock \emph{IEEE/CAA Journal of Automatica Sinica}, 10\penalty0
  (6):\penalty0 1361--1387, 2023.

\bibitem[Chung et~al.(2014)Chung, Gulcehre, Cho, and
  Bengio]{chung2014empirical}
Junyoung Chung, Caglar Gulcehre, KyungHyun Cho, and Yoshua Bengio.
\newblock Empirical evaluation of gated recurrent neural networks on sequence
  modeling.
\newblock \emph{arXiv preprint arXiv:1412.3555}, 2014.

\bibitem[Curtain \& Zwart(2020)Curtain and Zwart]{curtain2020introduction}
Ruth Curtain and Hans Zwart.
\newblock \emph{Introduction to infinite-dimensional systems theory: a
  state-space approach}, volume~71.
\newblock Springer Nature, 2020.

\bibitem[Evans(2022)]{evans2022partial}
Lawrence Evans.
\newblock \emph{Partial differential equations}, volume~19.
\newblock American Mathematical Society, 2022.

\bibitem[Farchi et~al.(2021)Farchi, Laloyaux, Bonavita, and
  Bocquet]{farchi2021using}
Alban Farchi, Patrick Laloyaux, Massimo Bonavita, and Marc Bocquet.
\newblock Using machine learning to correct model error in data assimilation
  and forecast applications.
\newblock \emph{Quarterly Journal of the Royal Meteorological Society},
  147\penalty0 (739):\penalty0 3067--3084, 2021.

\bibitem[Frion et~al.(2023)Frion, Drumetz, Mura, Tochon, and
  Bey]{frion2023neuralKoopmanAssimilation}
Anthony Frion, Lucas Drumetz, Mauro~Dalla Mura, Guillaume Tochon, and
  Abdeldjalil A{\"\i}ssa~El Bey.
\newblock Neural {Koopman} prior for data assimilation.
\newblock \emph{arXiv preprint arXiv:2309.05317}, 2023.

\bibitem[Gerlach(2014)]{gerlach2014semigroups}
Moritz Gerlach.
\newblock \emph{Semigroups of kernel operators}.
\newblock PhD thesis, Universit{\"a}t Ulm, 2014.

\bibitem[Goswami et~al.(2022)Goswami, Bora, Yu, and
  Karniadakis]{goswami2022physics}
Somdatta Goswami, Aniruddha Bora, Yue Yu, and George~Em Karniadakis.
\newblock Physics-informed neural operators.
\newblock \emph{arXiv preprint arXiv:2207.05748}, 2022.

\bibitem[Guen \& Thome(2020)Guen and Thome]{guen2020disentangling}
Vincent~Le Guen and Nicolas Thome.
\newblock Disentangling physical dynamics from unknown factors for unsupervised
  video prediction.
\newblock In \emph{Proceedings of the IEEE/CVF Conference on Computer Vision
  and Pattern Recognition}, pp.\  11474--11484, 2020.

\bibitem[Guibas et~al.(2021)Guibas, Mardani, Li, Tao, Anandkumar, and
  Catanzaro]{guibas2021adaptive}
John Guibas, Morteza Mardani, Zongyi Li, Andrew Tao, Anima Anandkumar, and
  Bryan Catanzaro.
\newblock Adaptive {Fourier} neural operators: Efficient token mixers for
  transformers.
\newblock \emph{arXiv preprint arXiv:2111.13587}, 2021.

\bibitem[Gupta et~al.(2021)Gupta, Xiao, and
  Bogdan]{gupta2021multiwaveletOperatorLearning}
Gaurav Gupta, Xiongye Xiao, and Paul Bogdan.
\newblock Multiwavelet-based operator learning for differential equations.
\newblock \emph{Advances in neural information processing systems},
  34:\penalty0 24048--24062, 2021.

\bibitem[Hao et~al.(2023)Hao, Wang, Su, Ying, Dong, Liu, Cheng, Song, and
  Zhu]{hao2023gnot}
Zhongkai Hao, Zhengyi Wang, Hang Su, Chengyang Ying, Yinpeng Dong, Songming
  Liu, Ze~Cheng, Jian Song, and Jun Zhu.
\newblock Gnot: A general neural operator transformer for operator learning.
\newblock In \emph{International Conference on Machine Learning}, pp.\
  12556--12569. PMLR, 2023.

\bibitem[Jiang et~al.(2021)Jiang, Meinert, Jord{\~a}o, Weisser, Holgate, Lavin,
  L{\"u}tjens, Newman, Wainwright, Walker, et~al.]{jiang2021digital}
Peishi Jiang, Nis Meinert, Helga Jord{\~a}o, Constantin Weisser, Simon Holgate,
  Alexander Lavin, Bj{\"o}rn L{\"u}tjens, Dava Newman, Haruko Wainwright,
  Catherine Walker, et~al.
\newblock Digital twin earth--coasts: Developing a fast and physics-informed
  surrogate model for coastal floods via neural operators.
\newblock \emph{arXiv preprint arXiv:2110.07100}, 2021.

\bibitem[Jiang et~al.(2023)Jiang, Yang, Wang, Huang, Xue, Chakraborty, Chen,
  and Qian]{jiang2023efficient}
Peishi Jiang, Zhao Yang, Jiali Wang, Chenfu Huang, Pengfei Xue, TC~Chakraborty,
  Xingyuan Chen, and Yun Qian.
\newblock Efficient super-resolution of near-surface climate modeling using the
  {Fourier} neural operator.
\newblock \emph{Journal of Advances in Modeling Earth Systems}, 15\penalty0
  (7):\penalty0 e2023MS003800, 2023.

\bibitem[Kaltenbach et~al.(2023)Kaltenbach, Perdikaris, and
  Koutsourelakis]{kaltenbach2023semi}
Sebastian Kaltenbach, Paris Perdikaris, and Phaedon-Stelios Koutsourelakis.
\newblock Semi-supervised invertible neural operators for {Bayesian} inverse
  problems.
\newblock \emph{Computational Mechanics}, pp.\  1--20, 2023.

\bibitem[Kassam \& Trefethen(2005)Kassam and Trefethen]{kassam2005fourth}
Aly-Khan Kassam and Lloyd~N Trefethen.
\newblock Fourth-order time-stepping for stiff pdes.
\newblock \emph{SIAM Journal on Scientific Computing}, 26\penalty0
  (4):\penalty0 1214--1233, 2005.

\bibitem[Kidger et~al.(2020)Kidger, Morrill, Foster, and
  Lyons]{kidger2020neural}
Patrick Kidger, James Morrill, James Foster, and Terry Lyons.
\newblock Neural controlled differential equations for irregular time series.
\newblock \emph{Advances in Neural Information Processing Systems},
  33:\penalty0 6696--6707, 2020.

\bibitem[Kingma \& Ba(2014)Kingma and Ba]{kingma2014adam}
Diederik~P Kingma and Jimmy Ba.
\newblock Adam: A method for stochastic optimization.
\newblock \emph{arXiv preprint arXiv:1412.6980}, 2014.

\bibitem[Kovachki et~al.(2021)Kovachki, Li, Liu, Azizzadenesheli, Bhattacharya,
  Stuart, and Anandkumar]{kovachki2021neural}
Nikola Kovachki, Zongyi Li, Burigede Liu, Kamyar Azizzadenesheli, Kaushik
  Bhattacharya, Andrew Stuart, and Anima Anandkumar.
\newblock Neural operator: Learning maps between function spaces.
\newblock \emph{arXiv preprint arXiv:2108.08481}, 2021.

\bibitem[Krstic \& Smyshlyaev(2008)Krstic and Smyshlyaev]{krstic2008boundary}
Miroslav Krstic and Andrey Smyshlyaev.
\newblock \emph{Boundary control of PDEs: A course on backstepping designs}.
\newblock SIAM, 2008.

\bibitem[Kutz et~al.(2016)Kutz, Brunton, Brunton, and Proctor]{kutz2016dynamic}
Nathan Kutz, Steven Brunton, Bingni Brunton, and Joshua Proctor.
\newblock \emph{Dynamic mode decomposition: data-driven modeling of complex
  systems}.
\newblock SIAM, 2016.

\bibitem[Larson \& Bengzon(2013)Larson and Bengzon]{larson2013finite}
Mats Larson and Fredrik Bengzon.
\newblock \emph{The finite element method: theory, implementation, and
  applications}, volume~10.
\newblock Springer Science \& Business Media, 2013.

\bibitem[Levine \& Stuart(2022)Levine and Stuart]{levine2022framework}
Matthew Levine and Andrew Stuart.
\newblock A framework for machine learning of model error in dynamical systems.
\newblock \emph{Communications of the American Mathematical Society},
  2\penalty0 (07):\penalty0 283--344, 2022.

\bibitem[Levine(2023)]{levine2023machine}
Matthew~Emanuel Levine.
\newblock \emph{Machine Learning and Data Assimilation for Blending Incomplete
  Models and Noisy Data}.
\newblock PhD thesis, California Institute of Technology, 2023.

\bibitem[Li et~al.(2020{\natexlab{a}})Li, Kovachki, Azizzadenesheli, Liu,
  Bhattacharya, Stuart, and Anandkumar]{li2020fourier}
Zongyi Li, Nikola Kovachki, Kamyar Azizzadenesheli, Burigede Liu, Kaushik
  Bhattacharya, Andrew Stuart, and Anima Anandkumar.
\newblock Fourier neural operator for parametric partial differential
  equations.
\newblock \emph{arXiv preprint arXiv:2010.08895}, 2020{\natexlab{a}}.

\bibitem[Li et~al.(2020{\natexlab{b}})Li, Kovachki, Azizzadenesheli, Liu,
  Bhattacharya, Stuart, and Anandkumar]{li2020neural}
Zongyi Li, Nikola Kovachki, Kamyar Azizzadenesheli, Burigede Liu, Kaushik
  Bhattacharya, Andrew Stuart, and Anima Anandkumar.
\newblock Neural operator: Graph kernel network for partial differential
  equations.
\newblock \emph{arXiv preprint arXiv:2003.03485}, 2020{\natexlab{b}}.

\bibitem[Li et~al.(2022)Li, Liu-Schiaffini, Kovachki, Azizzadenesheli, Liu,
  Bhattacharya, Stuart, and Anandkumar]{li2022learning}
Zongyi Li, Miguel Liu-Schiaffini, Nikola Kovachki, Kamyar Azizzadenesheli,
  Burigede Liu, Kaushik Bhattacharya, Andrew Stuart, and Anima Anandkumar.
\newblock Learning chaotic dynamics in dissipative systems.
\newblock \emph{Advances in Neural Information Processing Systems},
  35:\penalty0 16768--16781, 2022.

\bibitem[Lu et~al.(2019)Lu, Jin, and Karniadakis]{lu2019deeponet}
Lu~Lu, Pengzhan Jin, and George~Em Karniadakis.
\newblock Deeponet: Learning nonlinear operators for identifying differential
  equations based on the universal approximation theorem of operators.
\newblock \emph{arXiv preprint arXiv:1910.03193}, 2019.

\bibitem[Magnani et~al.(2022)Magnani, Kr{\"a}mer, Eschenhagen, Rosasco, and
  Hennig]{magnani2022approximate}
Emilia Magnani, Nicholas Kr{\"a}mer, Runa Eschenhagen, Lorenzo Rosasco, and
  Philipp Hennig.
\newblock Approximate {Bayesian} neural operators: Uncertainty quantification
  for parametric {PDEs}.
\newblock \emph{arXiv preprint arXiv:2208.01565}, 2022.

\bibitem[Morrill et~al.(2021)Morrill, Salvi, Kidger, and
  Foster]{morrill2021neural}
James Morrill, Cristopher Salvi, Patrick Kidger, and James Foster.
\newblock Neural rough differential equations for long time series.
\newblock In \emph{International Conference on Machine Learning}, pp.\
  7829--7838. PMLR, 2021.

\bibitem[Pathak et~al.(2022)Pathak, Subramanian, Harrington, Raja,
  Chattopadhyay, Mardani, Kurth, Hall, Li, Azizzadenesheli,
  et~al.]{pathak2022fourcastnet}
Jaideep Pathak, Shashank Subramanian, Peter Harrington, Sanjeev Raja, Ashesh
  Chattopadhyay, Morteza Mardani, Thorsten Kurth, David Hall, Zongyi Li, Kamyar
  Azizzadenesheli, et~al.
\newblock Fourcastnet: A global data-driven high-resolution weather model using
  adaptive {Fourier} neural operators.
\newblock \emph{arXiv preprint arXiv:2202.11214}, 2022.

\bibitem[Rahman et~al.(2023)Rahman, Ross, and Azizzadenesheli]{rahman2023uno}
Md~Ashiqur Rahman, Zachary Ross, and Kamyar Azizzadenesheli.
\newblock U-{NO}: {U}-shaped neural operators.
\newblock \emph{Transactions on Machine Learning Research}, 2023.
\newblock ISSN 2835-8856.
\newblock URL \url{https://openreview.net/forum?id=j3oQF9coJd}.

\bibitem[Rankin(1993)]{rankin1993semilinear}
Samuel~M Rankin.
\newblock Semilinear evolution equations in {Banach} spaces with application to
  parabolic partial differential equations.
\newblock \emph{Transactions of the American Mathematical Society},
  336\penalty0 (2):\penalty0 523--535, 1993.

\bibitem[Revach et~al.(2022)Revach, Shlezinger, Ni, Escoriza, Van~Sloun, and
  Eldar]{revach2022kalmannet}
Guy Revach, Nir Shlezinger, Xiaoyong Ni, Adria~Lopez Escoriza, Ruud Van~Sloun,
  and Yonina Eldar.
\newblock {KalmanNet}: Neural network aided {Kalman} filtering for partially
  known dynamics.
\newblock \emph{IEEE Transactions on Signal Processing}, 70:\penalty0
  1532--1547, 2022.

\bibitem[Rotman et~al.(2023)Rotman, Dekel, Ber, Wolf, and Oz]{rotman2023semi}
Michael Rotman, Amit Dekel, Ran~Ilan Ber, Lior Wolf, and Yaron Oz.
\newblock Semi-supervised learning of partial differential operators and
  dynamical flows.
\newblock In \emph{Uncertainty in Artificial Intelligence}, pp.\  1785--1794.
  PMLR, 2023.

\bibitem[Salvi et~al.(2022)Salvi, Lemercier, and Gerasimovics]{salvi2022neural}
Cristopher Salvi, Maud Lemercier, and Andris Gerasimovics.
\newblock Neural stochastic {PDEs}: Resolution-invariant learning of continuous
  spatiotemporal dynamics.
\newblock \emph{Advances in Neural Information Processing Systems},
  35:\penalty0 1333--1344, 2022.

\bibitem[S{\"a}rkk{\"a} \& Svensson(2023)S{\"a}rkk{\"a} and
  Svensson]{sarkka2023bayesian}
Simo S{\"a}rkk{\"a} and Lennart Svensson.
\newblock \emph{Bayesian filtering and smoothing}, volume~17.
\newblock Cambridge university press, 2023.

\bibitem[Shi et~al.(2015)Shi, Chen, Wang, Yeung, Wong, and
  Woo]{shi2015convolutional}
Xingjian Shi, Zhourong Chen, Hao Wang, Dit-Yan Yeung, Wai-Kin Wong, and
  Wang-chun Woo.
\newblock Convolutional {LSTM} network: A machine learning approach for
  precipitation nowcasting.
\newblock \emph{Advances in neural information processing systems}, 28, 2015.

\bibitem[Singh et~al.(2021)Singh, Westlin, Eisenbarth, Losin, Andrews-Hanna,
  Wager, Satpute, Barrett, Brooks, and Erdogmus]{singh2021variation}
Ashutosh Singh, Christiana Westlin, Hedwig Eisenbarth, Elizabeth A~Reynolds
  Losin, Jessica~R Andrews-Hanna, Tor~D Wager, Ajay~B Satpute, Lisa~Feldman
  Barrett, Dana~H Brooks, and Deniz Erdogmus.
\newblock Variation is the norm: Brain state dynamics evoked by emotional video
  clips.
\newblock In \emph{2021 43rd Annual International Conference of the IEEE
  Engineering in Medicine \& Biology Society (EMBC)}, pp.\  6003--6007. IEEE,
  2021.

\bibitem[Smith et~al.(1962)Smith, Schmidt, and McGee]{smith1962application}
Gerald Smith, Stanley Schmidt, and Leonard McGee.
\newblock \emph{Application of statistical filter theory to the optimal
  estimation of position and velocity on board a circumlunar vehicle}, volume
  135.
\newblock National Aeronautics and Space Administration, 1962.

\bibitem[Sonner(2022)]{sonnerintroduction}
Stefanie Sonner.
\newblock An introduction to partial differential equations.
\newblock \emph{Radboud University, Nijmegen, IMAPP - Mathematics}, 2022.

\bibitem[Temam(1982)]{temam1982behaviourSemilinear_t0}
Roger Temam.
\newblock Behaviour at time t=0 of the solutions of semi-linear evolution
  equations.
\newblock \emph{Journal of Differential Equations}, 43\penalty0 (1):\penalty0
  73--92, 1982.

\bibitem[Thodi et~al.(2023)Thodi, Ambadipudi, and Jabari]{thodi2023fourier}
Bilal~Thonnam Thodi, Sai Venkata~Ramana Ambadipudi, and Saif~Eddin Jabari.
\newblock Fourier neural operator for learning solutions to macroscopic traffic
  flow models: Application to the forward and inverse problems.
\newblock \emph{arXiv preprint arXiv:2308.07051}, 2023.

\bibitem[Wazwaz(2010)]{wazwaz2010partial}
Abdul-Majid Wazwaz.
\newblock \emph{Partial differential equations and solitary waves theory}.
\newblock Springer Science \& Business Media, 2010.

\bibitem[Weikmann et~al.(2021)Weikmann, Paris, and
  Bruzzone]{weikmann2021timesen2crop}
Giulio Weikmann, Claudia Paris, and Lorenzo Bruzzone.
\newblock Timesen2crop: A million labeled samples dataset of {Sentinel} 2 image
  time series for crop-type classification.
\newblock \emph{IEEE Journal of Selected Topics in Applied Earth Observations
  and Remote Sensing}, 14:\penalty0 4699--4708, 2021.

\bibitem[Weissler(1979)]{weissler1979semilinear}
Fred Weissler.
\newblock Semilinear evolution equations in {Banach} spaces.
\newblock \emph{Journal of Functional Analysis}, 32\penalty0 (3):\penalty0
  277--296, 1979.

\bibitem[Wen et~al.(2022)Wen, Li, Azizzadenesheli, Anandkumar, and
  Benson]{wen2022u}
Gege Wen, Zongyi Li, Kamyar Azizzadenesheli, Anima Anandkumar, and Sally~M
  Benson.
\newblock {U-FNO—An} enhanced fourier neural operator-based deep-learning
  model for multiphase flow.
\newblock \emph{Advances in Water Resources}, 163:\penalty0 104180, 2022.

\end{thebibliography}
\bibliographystyle{iclr2024_conference}

\appendix
\section*{Appendix}

\begin{color}{blue}
\section{Derivation of equation (12)}
\label{app:derivations_equation12}

    First, let us note that as explained in \citep{afshar2022extended} equation~(11) is a solution to the observer system in equation (10) when $K(t)$ is strongly continuous, due to proposition~2.4 in \citep{afshar2022extended}. 
    To improve clarity, we have made precise the need for the strong continuity of $K(t)$ for the solution in equation (11) to be valid right before equation (11) in the revised paper.

    Equation (12) consists of a time-discretization of equation (11). It can be derived by using the semigroup property of $T(t)$ \citep{curtain2020introduction}, which satisfies 
    \[T(t+t')=T(t)T(t')\]
    for any pair $t,t'$. This way, identifying $t$ in equation (11) with the discrete time $t_k$ we can express $\hat{z}(t_k)$ as:
    {\small \begin{align}
        \hat{z}(t_k) ={}& T(t_k)\hat{z}_0 + \int_{0}^{t_k}T(t_k-s)\Big[G(\hat{z}(s),s) + K(s)[y(s) - C\hat{z}(s)]\Big]ds
        \\
        ={}&  T(t_k-t_{k-1}+t_{k-1})\hat{z}_0 + \int_{0}^{t_{k-1}}T(t_k-t_{k-1}+t_{k-1}-s)\Big[G(\hat{z}(s),s) + K(s)[y(s) - C\hat{z}(s)]\Big]ds
        \nonumber\\ &
        + \int_{t_{k-1}}^{t_k}T(t_k-s)\Big[G(\hat{z}(s),s) + K(s)[y(s) - C\hat{z}(s)]\Big]ds
        \\
        ={}& T(t_k-t_{k-1})T(t_{k-1})\hat{z}_0 + T(t_k-t_{k-1})\int_{0}^{t_{k-1}}T(t_{k-1}-s)\Big[G(\hat{z}(s),s) + K(s)[y(s) - C\hat{z}(s)]\Big]ds
        \nonumber\\ &
        + \int_{t_{k-1}}^{t_k}T(t_k-s)\Big[G(\hat{z}(s),s) + K(s)[y(s) - C\hat{z}(s)]\Big]ds
        \\
        ={}& T(t_k-t_{k-1}) \underbrace{\bigg[T(t_{k-1})\hat{z}_0 + \int_{0}^{t_{k-1}}T(t_{k-1}-s)\Big[G(\hat{z}(s),s) + K(s)[y(s) - C\hat{z}(s)]\Big]ds\bigg]}_{=\hat{z}(t_{k-1})}
        \nonumber \\ &
        + \int_{t_{k-1}}^{t_k}T(t_k-s)G(\hat{z}(s),s)ds + \int_{t_{k-1}}^{t_k}T(t_k-s)K(s)[y(s) - C\hat{z}(s)]ds \,,
    \end{align}}
    where we used the linearity of $T(t)$. This is now in the same form as equation (12).

    \end{color}

\section{Implementation details and the training procedure}
\label{app:implementationdetails}

For all the experiments, for the training of NODA we optimized the loss function $\gJ(\phi)$ using the Adam~\citep{kingma2014adam} optimizer, using a learning rate of $10^{-4}$ with a multiplicative learning rate scheduler which decays 0.5 every 50 epochs. We train the model for a total of 300 epochs, using a batch size of 32. The weighting parameter $\lambda$ was fixed as 0.5.
The backbone of NODA uses an FNO block for $\gW$. For details on the FNO architecture, please see \citep{li2020fourier}. The total number of parameters in the model was comparable to that used in the original FNO work, with a width of 64 and 20 frequencies per channel. The increase in learnable parameters for NODA compared to the FNO only depends on the choices made for the additional parameters \{$E$, $W_y$, $W_z$, $b$\} and possibly $\hat{C}^*$. In the current setting, $W_y$, $W_z$ are single linear layers, $b$ is a bias, and we choose $\hat{C}^*$ to be the identity matrix since $C$ was known a priori and equal to the identity operator in the experiments. To parametrize $E$ in a way that is flexible enough to approximate the measurement process for different data acquisition models, we considered a 2-layer fully connected neural network with ReLU activation function. Thus, in the current setting of NODA, its additional parameters 
constitute only light increase to the number of learnable parameters in the base FNO architecture. Hence, the training time of NODA remains comparable to the original FNO (2D+time). 
For the experiments on the KS equation, we chose a 1D FNO as our base model, composed of four Fourier layers, whereas for the experiments on the NS equation, we considered a 2D FNO, also with four Fourier layers.
To implement the MNO~\citep{li2022learning}, the FNO~\citep{li2020fourier} the Conv-LSTM~\citep{shi2015convolutional}, and the LSTM~\citep{chung2014empirical} baselines, we considered the setup described in the original papers.

For the NS equation, we train NODA using the first 500 time samples of the 180 trajectories contained in the training dataset. We choose $t_H^{\rm train}=300$ for training of the model on the NS dataset. This allows the model to use both the prediction and the correction terms of the loss function in the first 300 samples, while only using the prediction part for the remaining 200 snapshots. As for the KS dataset, we selected $t_H^{\rm train}$ as 500. We trained NODA on 1000 trajectories contained in the training dataset, each containing a total of 800 snapshots.  
We generate the ground truth trajectories for the KS equation using the exponential time-differencing fourth-order Runge-Kutta method from \citep{kassam2005fourth}. To generate the trajectories for the NS equation, we used the Crank–Nicolson scheme as described in~\citep{li2020fourier}.

All the experiments were performed on computation nodes with Intel Xeon Gold 6132 CPUs and Nvidia Tesla V100 GPUs. Codes used to implement NODA can be found at: \url{https://github.com/singh17ashu/NODA-Neural-Operator-with-Data-Assimilation}


\section{Results for the Korteweg-de Vries (KdV) equation} \label{app:KSresults}

\textbf{Korteweg-de Vries (KdV) equation:} The KdV PDE is given by 
\begin{align}
    \cblue{\frac{\partial z}{\partial t} = -z\frac{\partial z}{\partial x}-\frac{\partial^3 z}{\partial^3 x}\,,}
\end{align}
with $z(t) \in [0,J]$, where $J=128$, and $F$ is a nonlinear operator. A periodic boundary condition is considered for this equation, and initial conditions $z(0)=z_0$. \cblue{This is a third order semilinear PDE~\citep{sonnerintroduction}.} For more details on this PDE, please refer to~\citep{wazwaz2010partial}. This equation represents the propagation of waves on a fluid surface when subjected to a perturbation. 
We generated 1200 trajectories from random initializations $z_0$ sampled as described in~\citep{brandstetter2022lie}, containing $t_f\in\{100,150,200\}$ seconds each, with a timestep of $h=0.5$ seconds. The first 1000 trajectories were used for training NODA and the competing methods, while the remaining 200 were used for evaluating their performance. The resolution was fixed at 128 samples.
To test NODA we used a warm-up period of $t_H=40$ seconds.

\textbf{Benchmark Methods:} For the KdV equation, we compare NODA with the \textbf{FNO}~\citep{li2020fourier} and with the \textbf{GRU}~\citep{chung2014empirical} methods.

\begin{figure*}[tbh]
    \centering
    \includegraphics[width=0.7\linewidth]{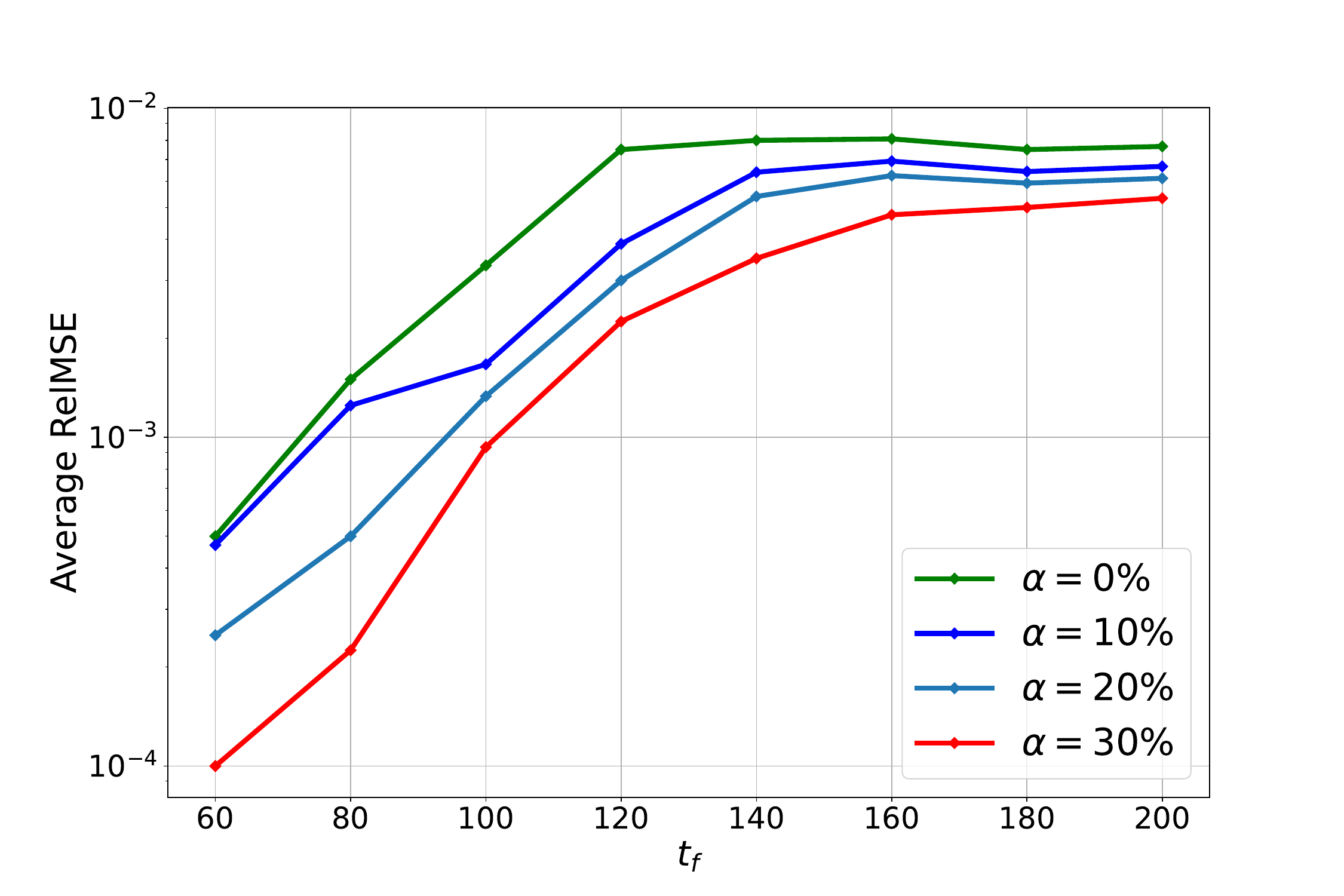}
    \vspace{-0.3cm}
    \caption{Average RelMSE as function of $t_f$ for different values of $\alpha$ for the KdV equation, where a warm-up period of $t_H=40$ being used for NODA.
    }
    \label{fig:averagerelRMSE_tf_KdV}
\end{figure*}

The average RelMSE results of all methods for prediction ($\alpha=0\%$) are shown in Table~\ref{tab:prediction_KDV}, where one can see a clear improvement on the performance of NODA when compared to the benchmark algorithms.
Regarding the data assimilation performance, Figure~\ref{fig:averagerelRMSE_tf_KdV} shows the Average RelMSE for different values of $t_f$ and for different values of $\alpha$. The results corroborate the findings stated in the main manuscript with the amount of available data directly impacting NODA's performance. 

\begin{figure*}[tbh]
    \centering
    \includegraphics[width=0.9\linewidth]{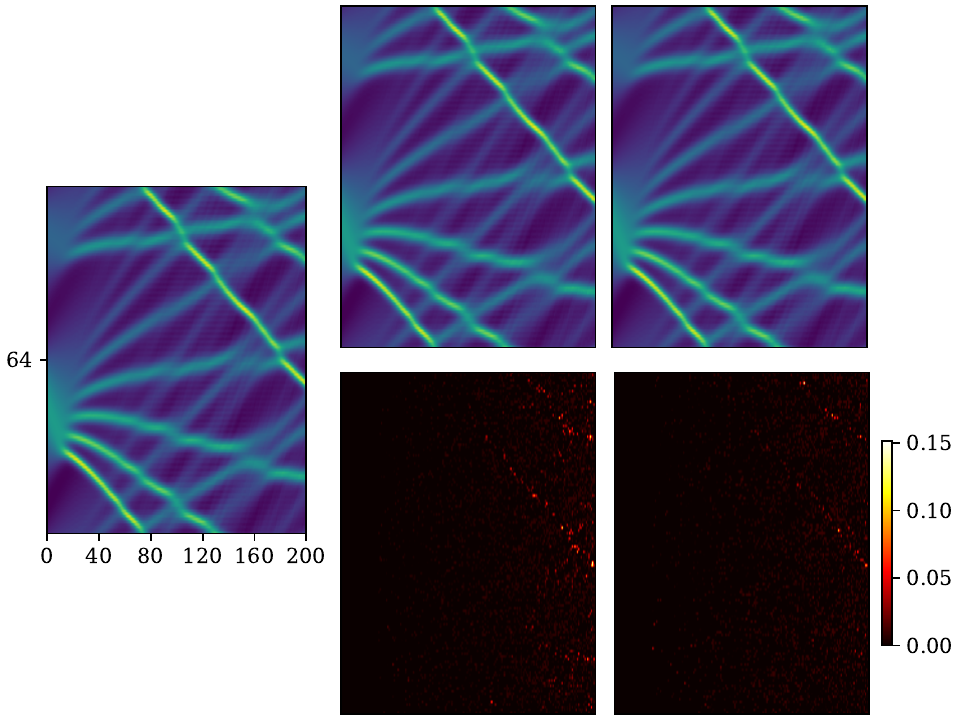}
    \vspace{-0.3cm}
    \caption{Sample of the predictionresults for the Korteweg-de Vries equation with $t_f=200$ seconds. 
    Ground truth trajectory (\textbf{Left Image}). Solution estimated by NODA with $\alpha=0\%$ (\textbf{Top left}). Solution estimated by NODA with $\alpha=30\%$ (\textbf{Top Right}). 
    Elementwise error plot for NODA's estimate without data assimilation (\textbf{Bottom Left}). Elementwise error plot for NODA's estimate with data assimilation (\textbf{Bottom Right}).} 
    \label{fig:appendix:KDV_error}
\end{figure*}

\begin{figure*}[tbh]
    \centering
    \includegraphics[width=0.9\linewidth]{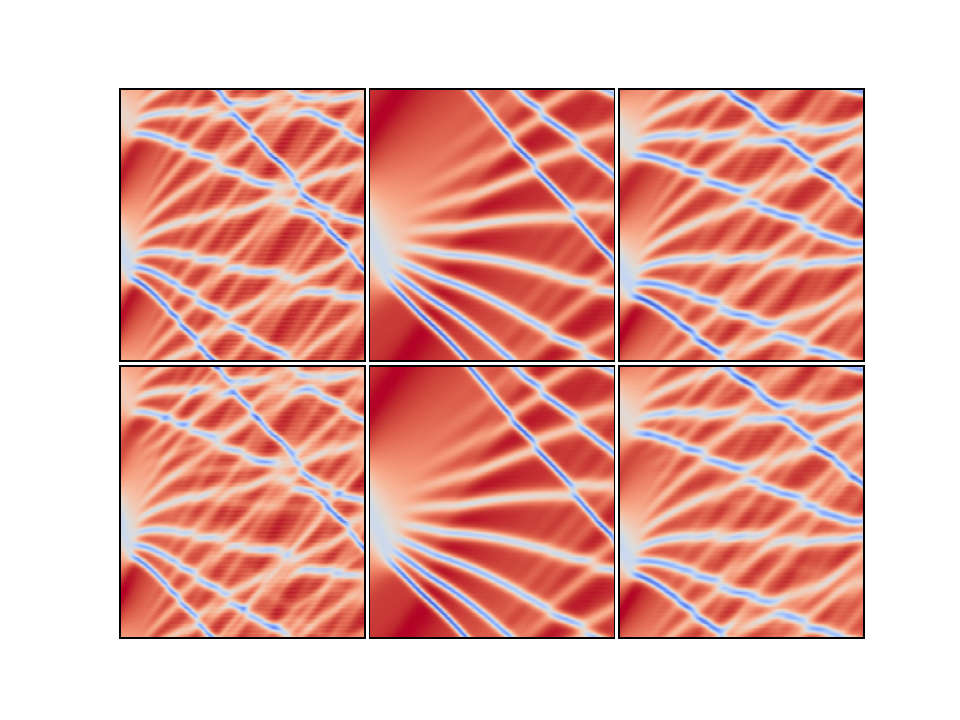}
    \vspace{-1.2cm}
    \caption{Sample of the prediction results for the Korteweg-de Vries equation with $t_f=200$ seconds. 
    Three different realisations of the ground truth trajectory generated from different initial conditions (\textbf{Top Row}). Corresponding trajectories predicted by NODA (\textbf{Bottom Row}).}
    \label{fig:appendix:KDV_est}
\end{figure*}

Plots containing samples of the ground truth solution, of NODA's estimation, and of elementwise estimation errors are shown in Figure~\ref{fig:appendix:KDV_error}, both with ($\alpha=30\%$) and without ($\alpha=0\%$) data assimilation. These results show that the estimation errors increase only moderately over time, and that data assimilation can significantly improve the estimation results.
Different samples of the ground truth trajectories and of the corresponding predictions by NODA for $\alpha=0\%$ are shown in Figure~\ref{fig:appendix:KDV_est}, where it is shown that NODA provides accurate predictions of the solution for different initial conditions.

\begin{table}[t]
\small
\centering
\renewcommand{\arraystretch}{1.2}
\caption{Averaged RelMSE \cblue{($\times10^3$)} for prediction ($\alpha=0\%$) on the Korteweg-de Vries equation as a function of the sequence length $t_f$ and of the SNR.} 
\smallskip
\begin{tabular}{l|rrr|rrr|rrr}
\hline

 SNR      & \multicolumn{3}{c|}{$\rm 20$ dB} &                                                      \multicolumn{3}{c|}{$\rm 30$ dB}   &      \multicolumn{3}{c}{$\rm \infty$}         \\ \cline{1-10}

 $t_f$                                            & 100          & 150               & 200          & 100          & 150                  & 200          & 100          & 150                  & 200          \\ \hline



FNO		&	21	&	129	&	164	&	17	&	95	&	134	&	13	&	79	&	112	\\	
MWNO		&	18	&	107	&	149	&	15	&	88	&	114	&	10	&	59	&	92	\\
GRU		&	207	&	296	&	421	&	190	&	219	&	258	&	122	&	178	&	254	\\	
NODA		&	\textbf{14}	&	\textbf{89}	&	\textbf{102}	&	\textbf{12}	&	\textbf{76}	&	\textbf{92}	&	\textbf{9}	&	\textbf{48}	&	\textbf{61}	\\	\hline

\end{tabular}
\label{tab:prediction_KDV}
\end{table}

\begin{color}{blue}
\section{Results for different measurement models}
In this section we evaluate the performance of NODA for two different choices of the discretized measurement operator $\hat{C}$. We present results for the NS example $(R_e = 40, t_f = 1000, \text{SNR}=30\text{dB})$, where the first choice of measurement operator is $\hat{C}=I$ (the identity operator, also considered in the main body of the paper), and $\hat{C}_{\rm rand}$, which is
a full matrix generated randomly, with each element sampled uniformly in the interval $\hat{c}_{ij} \in [0,1]$. In Table~\ref{tab:Cmatrix} we present the averaged RelMSE results for the two choices of the measurement operator and for different choices of $\alpha$. It can be seen that the example with the random operator $\hat{C}_{\rm rand}$ is noticeably more challenging, as evidenced by the consistently higher averaged RelMSE values. Nevertheless, the increase in the amount of data available for assimilation (i.e., $\alpha$) consistently increases the performance of NODA for both choices of measurement operator.

\begin{table}[tbh]
\small
\centering
\renewcommand{\arraystretch}{1.4}
\caption{Averaged RelMSE \cblue{($\times10^3$)} of NODA for the Navier-Stokes equation as a function of $\alpha$ for two different discretized measurement operators $\hat{C}$.}
\smallskip
\begin{tabular}{l|rrrr}
\hline
 $\alpha$  & 0\%  & 10\% & 20\% & 30\% \\ \hline
$\hat{C}=I$ &	26	&	18	&	13	&	9	\\
$\hat{C}_{\rm rand}$ & 42 & 39 & 33 & 25 \\ \hline
\end{tabular}
\label{tab:Cmatrix}
\end{table}

\section{Results for the Navier-Stokes equation with different vorticities}
\label{app:NS_more_vorticities}
In this section, we present results for NS equation with $Re \in \{500, 5000\}$, corresponding to flows with higher turbulence. We compare the results of NODA, MNO and FNO. For NODA we show the results for different values of $\alpha \in \{0\%, 10\%, 20\%, 30\% \}$, while for the remaining methods only $\alpha=0\%$ is shown as they do not perform data assimilation. For both choices of $Re$, we considered $\rm SNR=30dB$, $t_H = 50$ and $t_f = 500$. The averaged RelMSE results are shown in Table~\ref{tab:NS_higher_turbulence}.
From these results, it can be seen that when $\alpha=0\%$ (for prediction only), NODA outperforms FNO and has performance that is competitive with MNO, which provided the lowest average RelMSE values. However, as more data is used for assimilation (particularly for $\alpha\in\{20\%,30\%\}$), the average RelMSE of NODA decreases significantly, becoming lower than that those provided by both MNO and FNO. This highlights the benefits of data assimilation in the performance of NODA.
\begin{table}[t]
\small
\centering
\renewcommand{\arraystretch}{1.2}
\caption{Averaged RelMSE \cblue{($\times10^3$)} of NODA, MNO and FNO for the Navier-Stokes equation as a function of $\alpha$ for $Re\in\{500,5000\}$.}
\smallskip
\begin{tabular}{l|r|cccc}
\hline
                                           &  \backslashbox{$Re$}{$\alpha$}  & 0\%    & 10\%   & 20\%   & 30\% \\ \hline
\multicolumn{1}{c|}{\multirow{2}{*}{NODA}} & $500$  & 196 & 179 & 152 & 145 \\
\multicolumn{1}{c|}{}                      & $5000$ & 258 & 226 & 202 & 189 \\ \cline{1-6} 
\multirow{2}{*}{MNO}                       & $500$  & 183 & --    & --    & --    \\
                                           & $5000$ & 221 & --    & --    & --    \\ \cline{1-6} 
\multirow{2}{*}{FNO}                       & $500$  & 247 & --    & --    & --    \\
                                           & $5000$ & 310 & --    & --    & --    \\ \hline
\end{tabular}
\label{tab:NS_higher_turbulence}
\end{table}

\end{color}

\section{Additional results on the Navier-Stokes and Kuramoto-Sivashinsky equations}
\label{app:sec:moreresults}

\begin{figure*}[tbh]
    \centering
    \includegraphics[width=0.9\linewidth]{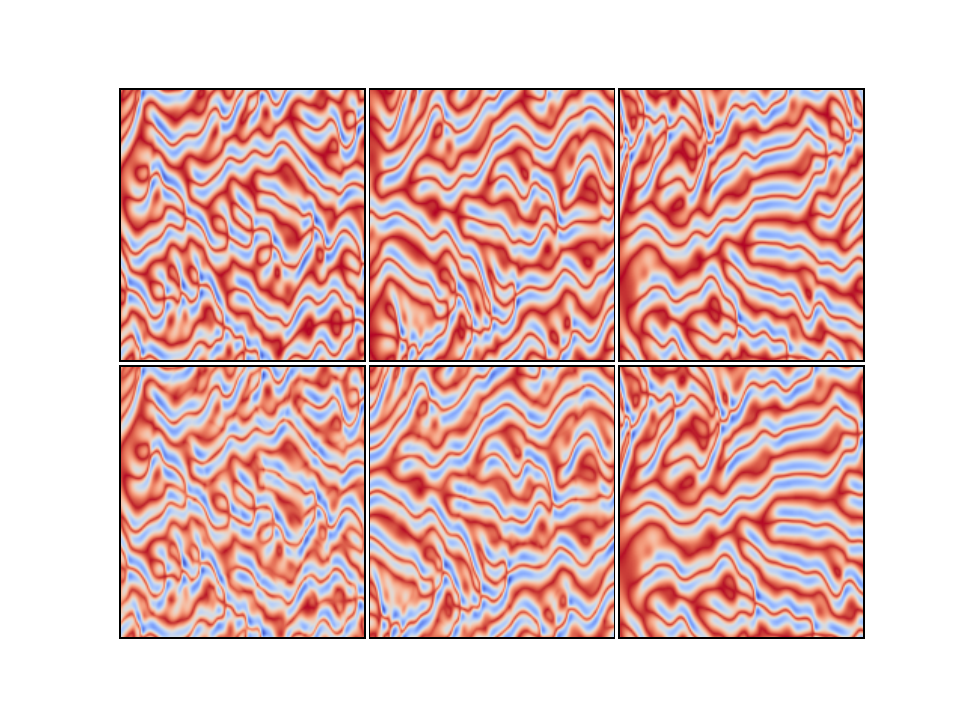}
    \vspace{-1.2cm}
    \caption{Different realizations of ground truth trajectories of the Kuramoto-Sivashinsky equation with $t_f=200$ seconds (\textbf{Top Row}), and the corresponding predictions obtained by NODA without data assimilation (\textbf{Bottom Row}).} 
    \label{fig:appendix:KS_est}
\end{figure*}

\begin{figure*}[tbh]
    \centering
    \includegraphics[width=0.9\linewidth]{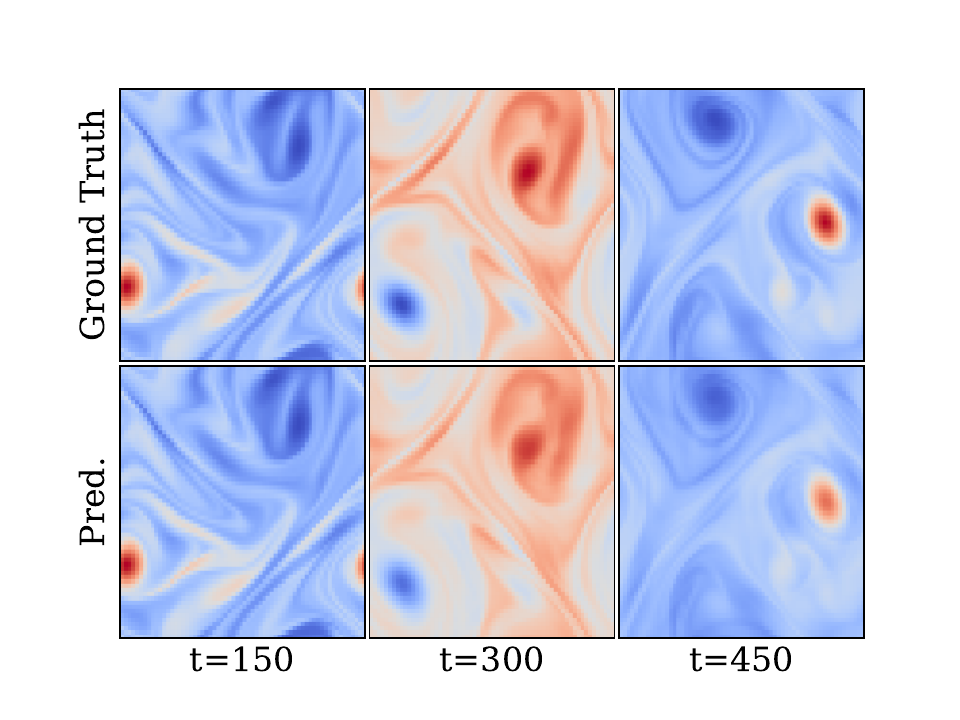}
    \vspace{-0.6cm}
    \caption{Sample of the ground truth sequence (\textbf{Top Row}) and predictions provided by NODA without data assimilation (\textbf{Bottom Row}) for the Navier-Stokes equation for $t\in\{150,300,450\}$.}
    \label{fig:appendix:NS_est}
\end{figure*}

In this section, we present complementary results for the KS and NS equations. These consist of visualizations of samples of the ground truth and reconstructed trajectories plots for predictions of NODA (using $\alpha=0\%$). These can be seen in Figures~\ref{fig:appendix:KS_est} and~\ref{fig:appendix:NS_est}. 

\end{document}